%% file: sn-article.tex
\documentclass[pdflatex,sn-mathphys-num]{sn-jnl}

\usepackage{caption}
\usepackage{graphicx}%
\usepackage{multirow}%
\usepackage{amsmath,amssymb,amsfonts}%
\usepackage{amsmath}%
\usepackage{mathrsfs}%
\usepackage[title]{appendix}%
\usepackage{xcolor}%
\usepackage{textcomp}%
\usepackage{manyfoot}%
\usepackage{booktabs}%
\usepackage{algorithm}%
\usepackage{algpseudocode}%
\usepackage{listings}%
\usepackage{algorithm}
\usepackage{multirow}
\usepackage{comment}
\DeclareMathOperator*{\argmin}{arg\,min}
\DeclareMathOperator*{\argmax}{arg\,max}

\usepackage{graphicx}
\usepackage{tikz}
\usepackage{url}
\usetikzlibrary{shapes.geometric, arrows, positioning}

\usepackage{fontawesome5} 

\tikzstyle{process} = [rectangle, rounded corners, draw=black, dashed, 
                        minimum width=3.2cm, minimum height=1.8cm, 
                        text centered, align=center]
                        
\tikzstyle{arrow} = [thick,->,>=stealth]

\theoremstyle{thmstyleone}%
\theoremstyle{thmstyletwo}%

\theoremstyle{thmstylethree}%

\begin{document}

\title[Article Title]{Filtering instances and rejecting predictions to obtain reliable models in healthcare}


\author*[1]{\fnm{Maria Gabriela} \sur{Valeriano }0000-0002-3631-156X}\email{mgv@unicamp.br}

\author[2]{\fnm{David} \sur{Kohan Marzagão } 0000-0001-8475-7913}\email{david.kohan@kcl.ac.uk}

\author[3]{\fnm{Alfredo} \sur{Montelongo } 0000-0002-6568-5448 }\email{alfredo.flores@telessauders.org.br}

\author[4]{\fnm{Carlos} \sur{Roberto Veiga Kiffer } 0000-0003-1122-0693 }\email{carlos.kiffer@unifesp.br}

\author[3]{\fnm{Natan} \sur{Katz }0000-0002-0659-7747}\email{nkatz@hcpa.edu.br}

\author[5]{\fnm{Ana Carolina} \sur{Lorena }0000-0002-6140-571X }\email{aclorena@ita.br}

\affil*[1]{\orgdiv{Instituto de Computação}, \orgname{Universidade Estadual de Campinas}, \orgaddress{\street{Av. Albert Einstein}, \city{Campinas}, \postcode{13083-889}, \state{São Paulo}, \country{Brazil}}}

\affil[2]{\orgdiv{Department of Informatics}, \orgname{King’s College London}, \orgaddress{\street{Aldwych G}, \city{London}, \postcode{WC2B 4BG}, \country{United Kingdom}}}

\affil[3]{\orgdiv{Núcleo de Telessaúde}, \orgname{Universidade Federal do Rio Grande do Sul}, \orgaddress{\street{Rua Dona Laura}, \city{Porto Alegre}, \postcode{10587}, \state{Rio Grande do Sul}, \country{Brazil}}}

\affil[4]{\orgdiv{Escola Paulista de Medicina}, \orgname{Universidade Federal de São Paulo}, \orgaddress{\street{Rua Botucatu}, \city{São Paulo}, \postcode{04044-020 }, \state{São Paulo}, \country{Brazil}}}

\affil[5]{\orgdiv{Divisão de Ciência da Computação}, \orgname{Instituto Tecnológico de Aeronáutica}, \orgaddress{\street{Praça Marechal Eduardo Gomes}, \city{São José dos Campos}, \postcode{12228-900}, \state{São Paulo}, \country{Brazil}}}


\abstract{Machine Learning (ML) models are widely used in high-stakes domains such as healthcare, where the reliability of predictions is critical. However, these models often fail to account for uncertainty, providing predictions even with low confidence. This work proposes a novel two-step data-centric approach to enhance the performance of ML models by improving data quality and filtering low-confidence predictions. The first step involves leveraging Instance Hardness (IH) to filter problematic instances during training, thereby refining the dataset. The second step introduces a confidence-based rejection mechanism during inference, ensuring that only reliable predictions are retained. We evaluate our approach using three real-world healthcare datasets, demonstrating its effectiveness at improving model reliability while balancing predictive performance and rejection rate. Additionally, we use alternative criteria—influence values for filtering and uncertainty for rejection—as baselines to evaluate the efficiency of the proposed method. The results demonstrate that integrating IH filtering with confidence-based rejection effectively enhances model performance while preserving a large proportion of instances. This approach provides a practical method for deploying ML systems in safety-critical applications.}


\maketitle

\section{Introduction}\label{sec1}

Machine learning (ML) models aim to capture relationships between input variables and a corresponding output feature from a training dataset. Once trained, these models are employed to make predictions on unseen instances. In critical applications such as healthcare, the stakes for these predictions are exceptionally high. A single misclassification, a missed diagnosis, or an incorrect treatment recommendation can have life-altering consequences for patients. Despite remarkable advances in predictive accuracy, ML models are not infallible. Conceptually, a model \(h\) is only an approximation of the true underlying function \(f\) relating the input space $\mathcal X$ to an output space $\mathcal Y$, and predictions are inherently uncertain, as they cannot be guaranteed to be correct \cite{hullermeier2021aleatoric}.

Discrepancies between \(h\) and \(f\) can arise due to various factors, including inconsistent or noisy data, overlapping classes, insufficient training samples, unexplored regions of the feature space, or incorrect modeling assumptions. Despite these discrepancies, traditional ML models assume that a prediction must always be provided, even when the likelihood of inaccuracy is high. This behavior is particularly concerning in safety-critical domains, where unreliable predictions can lead to severe consequences. ML systems must account for their limitations by quantifying uncertainty and recognizing when they lack sufficient information to make confident predictions, ensuring alignment with real-world safety requirements \cite{hendrickx2024machine}.

This work addresses these challenges by introducing a data-centric approach to enhance ML model performance and offer only confident predictions. Our framework leverages instance hardness— a measure of the likelihood that an instance will be misclassified \cite{smith2014instance}— as a central metric to guide decision-making. Using instance hardness, we refine datasets by removing problematic instances, thereby improving the model’s confidence in more straightforward test cases. We further complement this with a rejection mechanism during testing, ensuring that predictions with low confidence are filtered out.

We propose a novel two-step strategy: 

\begin{enumerate}
    \item \textit{Data Refinement:} Improving data quality by removing potential outliers and increasing class separability, which in turn boosts the model’s confidence and accuracy on less complex instances and;
    \item \textit{Confidence-Based Filtering:} During testing, we evaluate the model's confidence and reject low-confidence classifications. This ensures that only reliable predictions are retained.
\end{enumerate}
 
To validate our approach, we conduct three case studies using real-world healthcare datasets collected during routine operations within the Brazilian healthcare system. These datasets reflect the complexity and challenges of applying ML in critical domains. To complement these results, we also provide evaluations on benchmark datasets (Appendix \ref{apd_results}) that enable controlled comparisons and reinforce the generality of our approach.

\textcolor{black}{To compare the effectiveness of the proposed framework, we test an alternative criterion for each step. An Influence (IF) based metric was adopted as a baseline for instance hardness, while an uncertainty estimation provided a reference for comparing rejection based on confidence. In each case study, we demonstrate how our method affects model performance across four distinct scenarios, each testing a combination of adopted metrics for filtering and rejecting instances.  }

\textcolor{black}{Our findings illustrate that this adaptable strategy offers a robust and practical solution for enhancing the reliability of ML models in high-stakes applications. Besides that, the adoption of IH and the confidence rejection offered, in most cases, the preferred solution by balancing performance metrics and rejection rate.  }

The main contributions of this work can be summarized as follows:

\begin{itemize}
\item The integration of data filtering with a rejection strategy within a unified framework suitable for deployment in real-world scenarios.
\item The proposal of alternative criteria for filtering and rejection, employed as baselines to assess the effectiveness of the main approach.
\item A comprehensive analysis of three health-related datasets, two of which are publicly available. The framework implementation is also provided in our repository\footnote{https://github.com/gabivaleriano/RefineAI}.
\end{itemize}

Together, these contributions highlight both methodological innovation and practical relevance, particularly for high-stakes domains such as healthcare.

This paper is organized as follows: Section \ref{sec:dc} introduces the background and motivation for this research, framing it within a data-centric perspective. Section \ref{sec:methodology_rejection} presents the proposed framework. In Section \ref{sec:cases}, three case studies are discussed to demonstrate the framework’s performance on real-world data. Section \ref{sec:discussion} discusses the results and limitations. Finally, Section \ref{sec:conclusion} summarizes the main findings and conclusions. Appendix \ref{apd_results} presents additional case studies and results on benchmark datasets. Appendix \ref{apd:baselines} describes baselines used in the experimental study. Appendix \ref{apd_cost} details the minimization of the cost function of our method. And Appendix \ref{apd} presents information on the classifier used in the experiments.

\section{A data-centric paradigm} \label{sec:dc}

The success of ML systems is intrinsically tied to the quality of their training data. Biases, inconsistencies, and a lack of standardization can severely compromise model performance and reliability \cite{anik2021data}.
Regarding health data, the problem is particularly pronounced in regions such as Brazil, where non-uniform data-collection practices exacerbate existing issues \cite{rodriguez2020covid, valeriano2022let}. This context illustrates a broader challenge: even when large-scale data is available, a lack of quality and standardization undermines trustworthiness, underscoring the critical importance of focusing on data quality alongside quantity.

Despite its importance, data curation often receives insufficient attention due to practical constraints. Stakeholders may prioritize high-performance metrics over data quality, while data preparation processes can be costly and time-intensive \cite{seedat2022dc}. However, overlooking data quality can have significant consequences, especially in high-stakes domains such as healthcare, where unreliable predictions may lead to patient harm \cite{sambasivan2021everyone}. Addressing these gaps requires tailored solutions that enhance data quality while minimizing biases introduced during data collection and preprocessing. 

 This calls for a shift toward a data-centric paradigm, where the focus moves from refining model architectures to improving the quality and representativeness of the data itself \cite{zha2025data}. This approach recognizes robust and reliable datasets as the cornerstone of effective ML solutions. 
Emerging data-centric frameworks emphasize systematic evaluation and enhancement of datasets, ensuring alignment with the real-world scenarios they are intended to address \cite{seedat2022dataiq, seedat2022data}. Our approach embraces this paradigm by treating data as the foundational element of problem-solving, structuring solutions around the data rather than the model.

\subsection{Instance hardness: a data-centric measure}


Instance hardness (IH) is a powerful metric for assessing data quality, focusing on the classification difficulty of individual instances \cite{smith2014instance, seedat2024dissecting, liu2024instance}. Unlike cumulative metrics, IH provides granular insights into challenging instances, helping identify problematic data points that require further attention.

Defined empirically, IH can be quantified by the classification difficulty of an instance, based on the performance of multiple algorithms \cite{smith2014instance}. 
Formally, the instance hardness of a labeled instance \((\mathbf x_i, y_i)\) with respect to a set of algorithms \(\mathcal{L}\) is defined as:

\[
    IH_{\mathcal{L}}\big(\mathbf x_i, y_i \big) = 1 - \frac{1}{|\mathcal{L}|} \sum_{j=1}^{|\mathcal{L}|} p\big(y_i \mid \mathbf x_i, l_j(\mathcal D, \beta)\big),
\]

\noindent where \( p \) represents the probability of correct classification by algorithm \( l_j \) trained with hyperparameters \( \beta \) in a dataset \(\mathcal D \). This approach ensures that instances misclassified by most algorithms are identified as hard, while those correctly classified are marked as easy. By leveraging a diverse pool of classifiers with different inductive biases, this formulation captures consensus difficulty across models, providing robustness beyond what a single model can offer. 

\textcolor{black}{Other formulations to measure hardness level at an instance level have been proposed. To cite a few examples, GraNd (Gradient Normed) measures data hardness by estimating how much each training example influences the loss after a few steps of stochastic gradient descent \cite{paul2021deep}. The Data-IQ framework classifies instances into \textit{hard, ambiguous, and easy} based on their levels of aleatoric uncertainty and confidence. It can be calculated within any model trained in stages (iterations/epochs) \cite{seedat2022dataiq}. } 

Our motivation to adopt the formulation proposed by Smith and colleagues \cite{smith2014instance} lies in its aggregating nature, which combines the perspectives of multiple algorithms. Relying on a single model to remove instances may introduce bias. Within a data-centric paradigm, we therefore seek a solution that integrates information from different perspectives, reducing dependence on any individual model while capturing how data interacts with learning algorithms.

\textcolor{black}{Does averaging across models with different biases help capture patterns inherent to the data? Our experience with this formulation of IH suggests that it does, as it has proven valuable for developing data-centric solutions. In previous studies, we demonstrated the versatility of this approach by using it to build an explainability framework based on IH and meta-features \cite{valeriano2024explaining}, as well as to enhance preprocessing strategies, resulting in better-defined problems and improved predictive performance \cite{valeriano2024improving}.}

\textcolor{black}{To compare the efficacy of this IH formulation, we also implemented another hardness measure designed through a different perspective. Inspired by previous work \cite{Koh2017}, we leverage influence values to compute hardness levels from a single algorithm trained on different subsets of the training data, which is described in Appendix \ref{sec:IF}. This contrast allows us to evaluate whether consensus-based IH (across models) provides advantages over a single-model hardness perspective. 
}

The chosen formulation and the implemented baseline yield values in the interval [0, 1], enabling filtering by stage. In this way, it is possible to filter hard instances according to different thresholds \cite{smith2014instance}. However, filtering must be carefully calibrated, as excessive removal risks exclude valuable information, necessary for accurate predictions. 
We have shown in a previous work that filtering instances of the training data may harm models' performance on the test data \cite{Valeriano2025Beyond}.

This observation highlights a limitation of filtering alone: removing hard instances may discard rare yet informative cases, which is particularly critical in healthcare, where rare cases may hold significant clinical value \cite{seedat2022dataiq}. To address this, we advance beyond filtering by coupling it with a rejection mechanism at test time, ensuring both cleaner training data and safer predictions in deployment.


\subsection{Classification with a reject option} \label{sec:reject_option}

A reject option allows ML models to abstain from making predictions when the likelihood of error is high \cite{hendrickx2024machine}. By identifying regions of the feature space where predictions are uncertain, this mechanism improves model reliability and fosters trust in the system. In high-stakes settings like healthcare, it reduces the potential for harmful decisions and minimizes reliance on manual intervention.

The ability to defer decisions until sufficient confidence is achieved is a critical feature that enhances the model's utility in real-world applications. The integration of reject options into classification models has emerged as a pivotal advancement in ML \cite{hendrickx2024machine}. This approach allows nuanced decision-making that enhances both accuracy and reliability. 

One widely adopted strategy is the confidence-based rejector, where models output a confidence score alongside the predicted label. When this score falls below a predefined threshold, the model abstains from making a prediction \cite{traub2024overcoming}. This approach significantly reduces misclassifications, particularly in high-stakes environments where precision is paramount \cite{traub2024overcoming, hendrickx2024machine}. Our framework builds upon this strategy by combining confidence-based rejection at test time with data refinement at training time, ensuring that both learning and prediction stages contribute to reliability. 

Another important family of reject-option strategies is cost-sensitive formulations, which explicitly model rejection as an alternative decision with an associated risk \cite{bartlett2008classification}. In this setting, abstention is treated as a third possible outcome, alongside correct and incorrect classification, with its own predefined cost. The optimal strategy then minimizes the expected risk by balancing the trade-off between misclassification errors and rejection. This formulation provides a principled framework that can be adapted to different application requirements, since the cost of rejection can be tuned according to domain-specific tolerances. 

An alternative strategy is found in ensemble methods, where multiple models are combined and a reject decision is triggered when there is insufficient consensus among them \cite{varshney2011risk}. This ensemble-based rejection has been shown to improve robustness against noisy or ambiguous data \cite{xu2023efficient}. However, ensembles require additional computational resources and do not operate directly on the outputs of a single predictive model, making them less practical in certain real-world applications.

In our work, we focus on the confidence-based paradigm, enhanced by integrating instance filtering at the training stage. %
This confidence-based approach aligns naturally with our instance-hardness filtering strategy, allowing us to improve reliability without incurring the additional computational cost of ensembles or the complexity of modeling explicit rejection costs. By removing problematic instances and rejecting low-confidence predictions, our approach establishes a robust framework that increases model reliability while maintaining computational efficiency. 



Our approach integrates a dependent-rejector paradigm, in which the predictor and rejector operate as distinct yet complementary modules. The rejector independently evaluates predictions, refusing instances when the predictor's confidence falls below a specified threshold. Nevertheless, removing difficult instances from the training dataset has been shown to increase confidence in predictions \cite{seedat2022dataiq}. In this way, these stages, when combined, create a robust framework for enhancing model reliability.

 To compare with a more complex measure, we also implemented an uncertainty estimation extracted from a collection of ensembles. This estimation was inspired by previous work \cite{lakshminarayanan2017simple} and is described in more detail in Appendix \ref{sec:U}.

\subsection{Related work}

Comparing our approach to existing methods is challenging. To the best of our knowledge, this two-step formulation has not been proposed before. A point of reference can be found in the experiments conducted by Smith et al. \cite{smith2014instance} and Seedat et al. \cite{seedat2022data}, which also explored instance hardness. 

Smith et al. \cite{smith2014instance} explored the concept of instance hardness, filtering out difficult instances from various datasets and measuring model performance using cross-validation. The removal process was guided by different instance hardness thresholds. Their results showed that accuracy increased in 85\% of the datasets examined. They experimented with excluding instances above different IH levels and concluded that the optimal threshold varied across datasets and learning algorithms.

Similarly, Seedat et al. \cite{seedat2022data} proposed a method to quantify instance hardness by analyzing how individual samples behave during the training process of iterative algorithms. 
Their study also investigated the impact of filtering difficult instances by evaluating model performance through a cross-institutional setup, where training data from one hospital was tested on another. The findings suggested that reducing the proportion of difficult instances generally led to improved accuracy, though the effect was not uniform across all cases.

However, both studies have certain limitations. Their evaluations focus primarily on overall accuracy, which may obscure declines in performance for minority classes in imbalanced datasets. Our experience shows that removing hard instances can lead to varying behavior across evaluation metrics, highlighting the importance of using diverse metrics for a comprehensive analysis \cite{Valeriano2025Beyond}.  
Additionally, Smith et al. \cite{smith2014instance} did not assess how the models generalized to external data sources, whereas we observe that performance improvements from filtering hard instances do not always generalize \cite{Valeriano2025Beyond}.

\section{Methods} \label{sec:methodology_rejection}

This study aims to address the central question: \textit{Can filtering hard instances and rejecting low-confidence predictions significantly improve model performance and confidence on the accepted instances?} Formally, our approach can be defined as follows. 

Let \( \mathcal{D} = \{(\mathbf x_i, y_i)\}_{i=1}^{N} \) be the original dataset, where \( \mathbf x_i \in \mathcal{X} \) represents an instance and \( y_i \in \mathcal{Y} = \{0,1\} \) denotes its corresponding class label in a binary classification setting\footnote{The proposal can be easily adapted to multiclass and regression problems as we discuss in Section \ref{sec:discussion}.}. The training data is filtered to create a refined set:

\[
\mathcal{D}_f = \{(\mathbf x_i, y_i) \in \mathcal{D} \mid \text{IH}(\mathbf x_i,y_i) \leq T_f \},
\]

\noindent where \( \text{IH}(\mathbf x_i,y_i) \) represents the instance hardness score of instance \(\mathbf x_i \), and \( T_f \) is the filtering threshold. The classifier \( h: \mathcal{X} \to \mathcal{Y} \) is then trained on \( \mathcal{D}_f \).  
At inference time, the decision function \( g: \mathcal{X} \to \mathcal{Y} \cup \{\text{reject}\} \) decides whether to accept or reject a prediction based on a confidence score \( s(\mathbf x) \) produced by the classifier. 

Model confidence is estimated from the conditional distribution $P(Y|X)$, obtained via post-processing the predictor's output. In our case, we employ sigmoid calibration for binary classification tasks, to adjust the predicted probabilities \cite{boken2021appropriateness}. 
A logistic sigmoid function is applied to the output probabilities, mapping them to the range \([0,1]\) while preserving their ranking order. 

The calibrated probability obtained can be described as:

\[
\hat{p}(y \mid \mathbf x) = \sigma(a \cdot p(y \mid \mathbf x) + b),
\]

\noindent where \( \sigma(z) = \frac{1}{1 + e^{-z}} \) is the sigmoid function, and \( a, b \) are learned calibration parameters estimated by fitting a logistic regression model on the outputs of the uncalibrated classifier. Specifically, they are estimated by minimizing the negative log-likelihood between the calibrated probabilities and the true labels inside a five-fold cross-validation strategy on the training data. 

Once the calibrated probabilities are obtained, the confidence score is defined as the maximum class probability \cite{hendrickx2024machine}:

\[
s(\mathbf x) = \max_{y \in \mathcal{Y}} \hat{p}(y \mid \mathbf x),
\]


\noindent And the final decision function is: 
\[
g(\mathbf x) =
\begin{cases}
h(\mathbf x), & \text{if } s(\mathbf x) \geq T_r, \\
\text{reject}, & \text{otherwise}.
\end{cases}
\]

\noindent Where \( T_r \) is the rejection threshold applied during inference. 

This two-step process, filtering the training data with \( T_f \) and rejecting uncertain predictions with \( T_r \), aims to train the model on reliable samples and avoids making predictions in ambiguous cases. Our framework thus revolves around two thresholds: \(T_f\) for filtering hard instances during training and \(T_r\) for rejecting low-confidence predictions during inference. 

\subsection{Trade-offs in training models with a reject option}

Evaluating ML models with a rejection option introduces complexity, as it requires balancing the competing objectives of improving prediction accuracy while maintaining sufficient coverage \cite{hendrickx2024machine}. Higher performance often comes at the cost of rejecting more instances, thereby reducing the number of predictions. Therefore, determining suitable thresholds for filtering and rejecting instances is critical to achieving this balance.

ML models might display problematic behaviors at two extremes. On the one hand, a model may exhibit overconfidence when assigning predictions to incorrect classifications. This can mislead decision-makers, instilling undue trust in faulty outcomes and amplifying potential risks. On the other hand, models may limit their confidence to a narrow subset of instances, leaving the majority of cases with low-confidence predictions and diminishing their practical utility. These two extremes highlight the crucial need to strike a balance between performance and confidence in ML predictions.

Ideally, an ML model should achieve high predictive accuracy and high confidence. Adopting a classification model with a reject option allows retaining only high-confidence predictions, often increasing average model performance. This means excelling at making accurate predictions and reliably identifying instances where a confident decision cannot be made. However, setting an appropriate rejection threshold is challenging due to inherent trade-offs in model design. 

\begin{figure}[htbp!]
    \centering
    \includegraphics[scale=0.55]{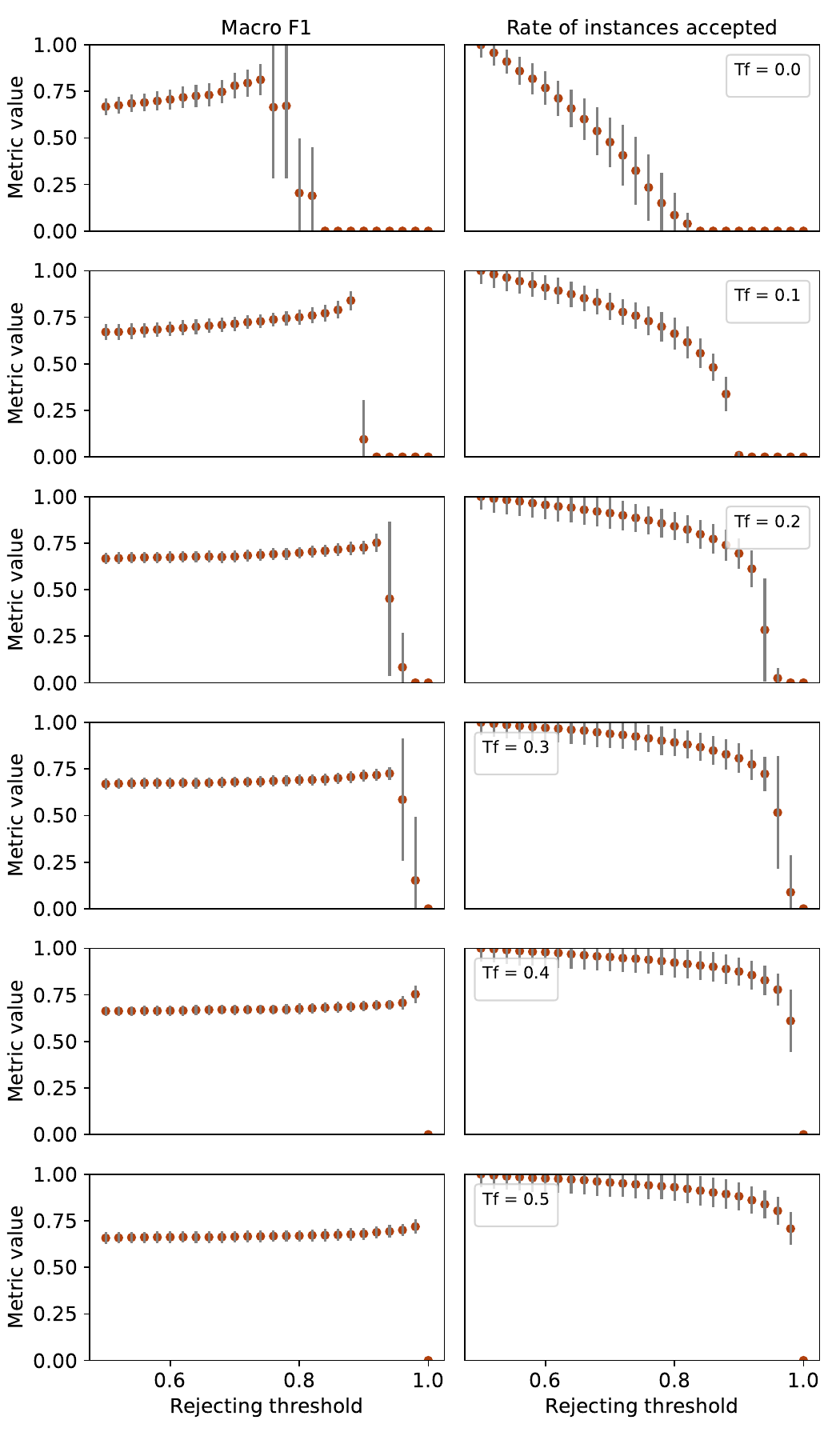} 
    \caption{\small
    Average macro-F1 and rate of accepted instances computed across five train-validation splits. The acceptance rate is relative to the size of the validation set. Models were trained using XGBoost with varying filtering threshold values (\(T_f\)) and evaluated across a range of rejection thresholds (\(T_r\)). Experiments were conducted using the \textit{severity\_hsl} dataset. As \(T_r\) increases, predictive performance improves for models trained with lower \(T_f\) values. Higher \(T_f\) values result in more stable performance metrics across different \(T_r\) levels.}
    \label{fig:example-figure} 
\end{figure}

Improving coverage—the volume of predictions made—involves encouraging generalization, which can lead to uncertain classifications for edge cases or novel scenarios being accepted. Conversely, prioritizing confidence and performance may yield overly conservative models that reject many predictions, potentially reducing their utility in real-world applications. 
In healthcare, this trade-off is particularly important, as even a few reliable predictions can substantially ease the burden on human experts.
To illustrate the relation between performance and rejection rate, we explore the effect of different threshold combinations in Figure \ref{fig:example-figure}.

In Figure \ref{fig:example-figure}, we conducted a series of experiments using the \textit{severity\_hsl} dataset, which will be further described and analyzed in Section \ref{dados:hsl}. Each row illustrates how the macro-F1 score and rate of acceptance evolve as the rejection threshold \(T_r\) increases. The acceptance rate is the proportion of accepted instances in the validation set relative to the total number of instances. The training and validation sets were created by splitting the training data into a 70:30 ratio. This process was repeated five times with different random seeds. The dots in the graphs represent the average results across these five experiments. The algorithm used in our experiments was XGBoost, with hyperparameters detailed in Appendix \ref{apd}.

The first row represents models trained on all instances, the second row removing 5\% of the most challenging instances, the third row removing 10\%, and so on. When \(T_f\) is low, the model is exposed to a broader range of instances, including more challenging ones. As a result, applying confidence-based rejection leads to greater performance fluctuations, which is reflected in the large standard deviation bars. Filtering out difficult instances in the training set results in a more robust model.

 As \(T_r\) increases, an initial improvement in predictive performance is observed. This effect is most pronounced in the first few rows, where models were trained on datasets filtered with lower \(T_f\) values. In these cases, models perform better when higher confidence levels are required for classification. However, increasing \(T_r\) also reduces the acceptance rate, as only the most confident predictions are retained. This creates a trade-off between prediction volume and performance. 

Without filtering, the highest confidence achieved remains below 0.7, as \(T_f\) values increase, the model's confidence rises. At the same time, when hard instances are filtered, performance metrics across different \(T_r\) values become more stable, though at intermediate performance levels. For instance, with \(T_f = 0.1\), mean performance nearly reaches 0.8 as \(T_r\) increases, whereas with \(T_f = 0.4\), performance remains more stable across all \(T_r\) values but around 0.7.

This exploration suggests that filtering with low \(T_f\) values (excluding fewer hard instances) leads to better performance when evaluated with higher \(T_r\) values. Retaining a diverse set of instances during training helps the model generalize more effectively when confidence-based filtering is applied at test time. Conversely, as \(T_f\) increases (filtering out more difficult instances), the models become more stable but may not achieve improved performance metrics.
This underscores the challenge of selecting a set of threshold values that balances high-confidence predictions, strong predictive performance, and minimal instance rejection.

\subsection{Evaluating models with a reject option} \label{sec:evaluating}

In our study, each \(T_f\) value generates a new dataset, which is then used to train a model subsequently evaluated across multiple rejection thresholds. This iterative process enables the identification of the thresholds that offer a suitable balance between performance and confidence. 
When comparing different models, one alternative for evaluating reject-option mechanisms is to adopt a fixed rejection rate, which can be set based on expert knowledge. Then, performance is assessed on the non-rejected instances, and the model with the highest performance can be selected \cite{hendrickx2024machine}. 
\textcolor{black}{Although it is possible to set a coverage rate in our framework,} we aim to retain as many predictions as possible, while avoiding both incorrect predictions and unnecessary rejections. 

Another method to evaluate prediction quality involves varying the rejection rate and plotting the results on an Accuracy-Rejection Curve (ARC) \cite{hendrickx2024machine}. Higher curves indicate better performance. ARCs can be constructed with accuracy or other relevant metrics, offering a comprehensive overview of model performance.
Based on the exploration conducted in the previous section, we concluded that this strategy is not ideal in our case. The highest ARC value would be attributed to models trained with \(T_f\) = 0.5. However, at this stage, the predictive performance remains stable at around 0.65 across most of the \(T_r\) values tested. This illustrates why, in our framework, the highest ARC does not necessarily indicate the best combination of filtering and rejection thresholds. 

In summary, three variables must be balanced: confidence level, number of instances rejected, and predictive performance across one or more metrics. 
This complexity makes the choice of thresholds a challenging task. To formalize threshold selection, we adopt a cost function that balances the three factors: confidence, performance, and rejection rate. We aim to find a combination of thresholds \( T_f \) and \( T_r \) that minimize the following cost function:

\[
\text{Cost} = w_p (1 - F_1) + w_r R_r + w_c (1 - C),
\]

\noindent where:

\begin{itemize}
    \item \( C \) is the {mean confidence} across the validation set.
    \item \( F_1 \) is the mean {model performance}, measured with the macro-\(F_1 \) score.
    \item \( R_r \) is the {rejection rate}, given by:
    
\small
\[
R_r  = 1 - \frac{\text{Number of Instances Classified}}{\text{Total Number of Instances}}.
\]
\normalsize
\end{itemize}

\vspace{0.5cm}

The minimization search can be formulated as: 
\[
(T_f, T_r) = \arg\min_{T_f, T_r} \text{(Cost)}.
\]



The macro-\( F_1 \) metric is chosen because it provides a balanced assessment of model performance across all classes, regardless of class distribution. Unlike the standard \( F_1 \) score, which may be dominated by the majority class in imbalanced datasets, macro-\( F_1 \) calculates the \( F_1 \) score for each class independently and averages them. This ensures that performance across underrepresented classes is given equal importance, which is particularly crucial in applications where balanced classification is essential. 
However, the performance metric is flexible and can be adapted to the deployment scenario.


The weights \( w_p \), \( w_r \), and \( w_c \) in the cost function determine the trade-offs between model performance, rejection rate, and confidence. These parameters allow flexibility in optimizing the selection thresholds by prioritizing one aspect over another.
In our experiments, we found that setting \( w_p = 4 \), \( w_r = 1 \), and \( w_c = 1 \) provides a reasonable starting point. This configuration prioritizes model performance while still considering rejection rate and confidence. 

Intuitively, this means that improving performance is four times more important than reducing the rejection rate or increasing confidence, underscoring the need to prioritize accuracy in healthcare decision support systems. This setup ensures that the model does not minimize cost solely by lowering the rejection rate. Instead, it enhances performance by strategically rejecting instances. The weight values can be adjusted based on the specific requirements of a given application. 

\begin{figure*}[htbp!]
    \centering
    \begin{minipage}{0.48\textwidth}
        \centering
        \includegraphics[width=\textwidth]{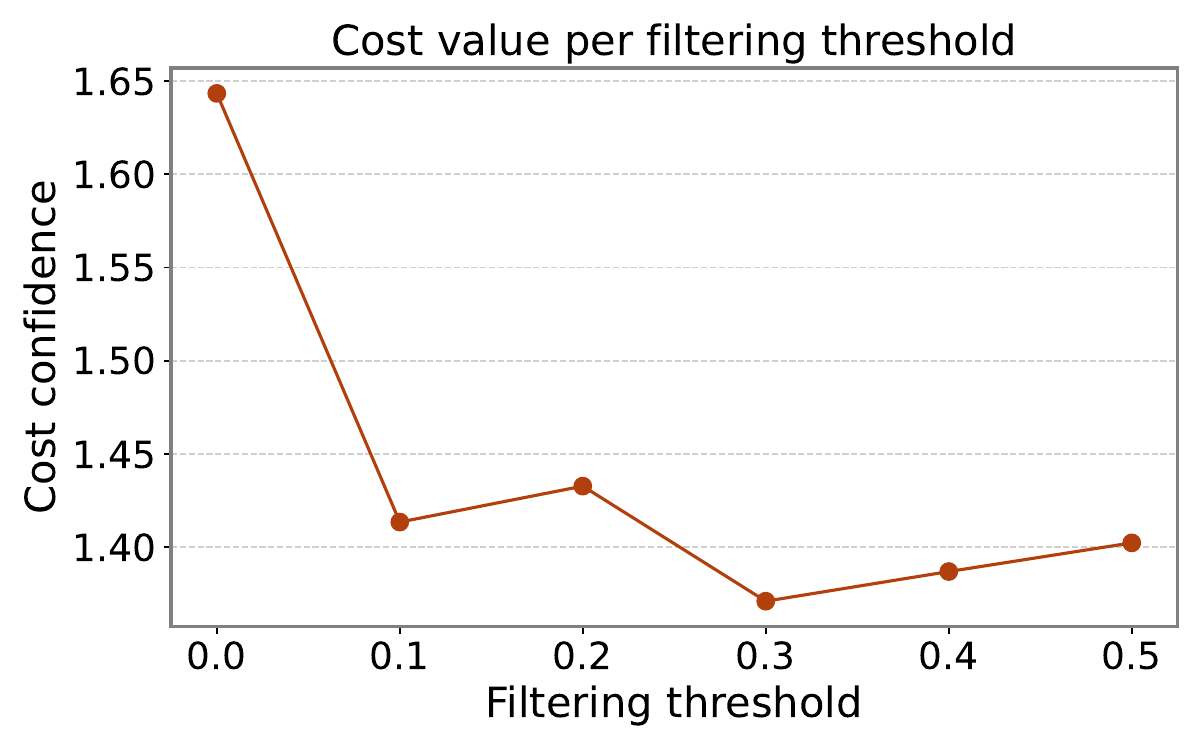} 
        \label{fig:first_plot}
    \end{minipage}
    \hfill
    \begin{minipage}{0.48\textwidth}
        \centering
        \includegraphics[width=\textwidth]{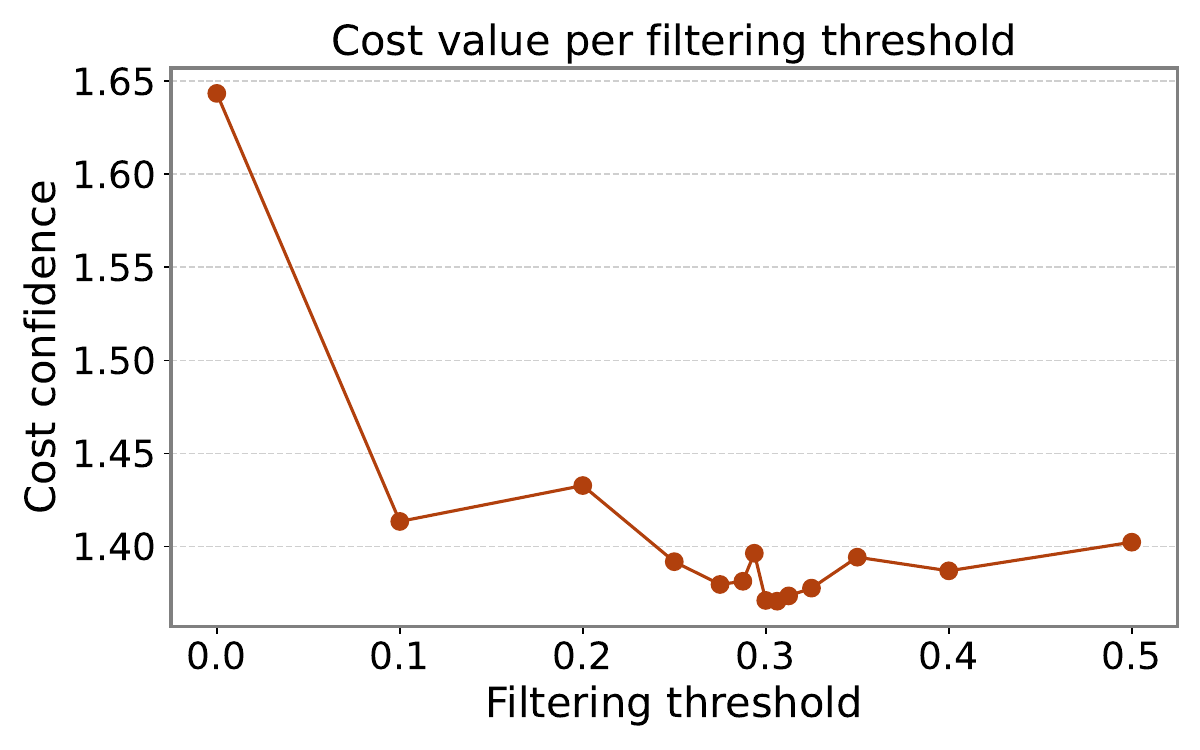} 
        \label{fig:second_plot}
    \end{minipage}
    
    \caption{\small Cost values per filtering threshold applied to the \textit{severity\_hsl} dataset. Instances were filtered based on IH values, and confidence-based rejection was applied. For each filtering threshold we selected the rejection threshold that minimizes the cost. Values are averages over five different train-validation splits.
    The first (left) plot presents the initial set of filtering thresholds tested, whereas the second plot (right) includes additional values identified using a heuristic approach.}
    \label{fig:comparison}
\end{figure*}

With the cost function in place, finding the best rejection threshold (\( T_r \)) for each model becomes straightforward. However, the challenge lies in selecting the filtering threshold (\( T_f \)). Figure
\ref{fig:comparison} (left) illustrates the cost for six different models trained on data filtered using varying \( T_f \) values. This investigation was conducted using the \textit{severity\_hsl} dataset, as described in Section \ref{dados:hsl}. 

Examining the behavior of the cost function reveals a pattern, though it is not always monotonic. Some fluctuations exist, indicating that the relationship between \( T_f \) values and the cost function is not consistently decreasing. We adopted a heuristic approach to select a suitable option, thereby avoiding the need to exhaustively test many \( T_f \) values. \textcolor{black}{In Appendix \ref{apd_cost}, we explore other methods to minimize the cost function and justify our decision. }

Our approach begins with a grid search, testing a predefined set of \( T_f \) values: 0.5, 0.6, 0.7, 0.8, 0.9, and 1.0. Rather than testing a large number of \( T_f \) values, which would be computationally expensive, this initial grid search provides a reasonable range of candidates. 
After the grid search, we apply a local search around the best-performing \( T_f \) value, in both directions, by progressively narrowing the range based on performance. This allows us to fine-tune the selection process without having to explore every potential value. 

The heuristic provides a balance between performance and computational efficiency, reducing the need for exhaustive testing while ensuring that we converge on a well-performing threshold. In the Appendix \ref{apd_cost}, we present the algorithm for the adopted heuristic. 
This strategy is not guaranteed to be optimal. However, as we explore the neighborhood of a \( T_f \) value, it is reasonable to expect slight changes in results and cost values, but not strong perturbations. In this way, the heuristic may be adopted without further complications. Figure \ref{fig:comparison}, on the right, presents all the  \( T_f \) explored through this approach and their corresponding cost values. 

\subsection{Framework description}

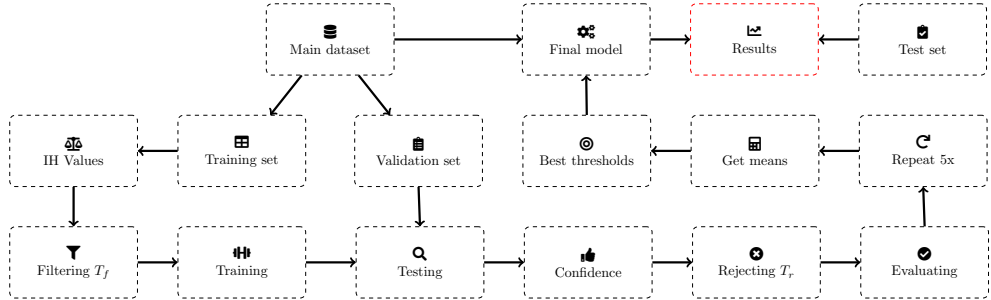
\begin{figure*}[htbp]
    \centering
    \resizebox{0.995\textwidth}{!}{ 
    \begin{tikzpicture}[node distance=2cm]

        \node (dataset) [process] {\faDatabase \\ Main dataset};
        \node (finalmodel) [process, right=3.2cm of dataset] {\faCogs \\ Final model};
        \node (results) [process, right=1cm of finalmodel, draw=red] {\faChartLine \\ Results};
        \node (testset) [process, right=1cm of results] {\faClipboardCheck \\ Test set};

        \draw [ultra thick, ->] (dataset) -- (finalmodel);
        \draw [ultra thick, ->] (finalmodel) -- (results);
        \draw [ultra thick, ->] (testset) -- (results);

        \node (train) [process, below left=1cm and -1cm of dataset] {\faTable \\ Training set};
        \node (ih) [process, left=1cm of train] {\faBalanceScale \\ IH Values};
        \node (validation) [process, below right=1cm and -1cm of dataset] {\faClipboardList \\ Validation set};
        \node (mincost) [process, right=1cm of validation] {\faBullseye \\ Best thresholds};
        \node (means) [process, right=1cm of mincost] {\faCalculator \\ Get means};
        \node (repeat) [process, right=1cm of means] {\faRedo \\ Repeat 5x};

        \draw [ultra thick, ->] (dataset) -- (train);
        \draw [ultra thick, ->] (dataset) -- (validation);
        \draw [ultra thick, ->] (means) -- (mincost);
        \draw [ultra thick, ->] (repeat) -- (means);
        \draw [ultra thick, ->] (train) -- (ih);

       
        \node (filtering) [process, below =1cm of ih] {\faFilter \\ Filtering \(T_f\)};
        \node (training) [process, right=1cm of filtering] {\faDumbbell \\ Training};
        \node (testing) [process, right=1.25cm of training] {\faSearch \\ Testing};
        \node (confidence) [process, right=1cm of testing] {\faThumbsUp \\ Confidence};
        \node (rejecting) [process, right=1cm of confidence] {\faTimesCircle \\ Rejecting \(T_r\)};
        \node (evaluating) [process, right=1cm of rejecting] {\faCheckCircle \\ Evaluating};

        \draw [ultra thick, ->] (ih) -- (filtering);
        \draw [ultra thick, ->] (filtering) -- (training);
        \draw [ultra thick, ->] (training) -- (testing);
        \draw [ultra thick, ->] (testing) -- (confidence);
        \draw [ultra thick, ->] (confidence) -- (rejecting);
        \draw [ultra thick, ->] (rejecting) -- (evaluating);
        \draw [ultra thick, ->] (evaluating.north) -- (repeat.south);

        \draw [ultra thick, ->] (mincost.north) -- (finalmodel.south);
        \draw [ultra thick, ->] (validation.south) -- (testing.north);

    \end{tikzpicture}
    }
   \caption{\small Framework used in this study. The main dataset is divided into training and validation sets. IH values are estimated from the training set, and instances are filtered using different thresholds (\( T_f \)). The filtered data is used to train a classifier, which is evaluated on the validation set at various rejection thresholds (\( T_r \)) based on classifier confidence. This process is repeated 5 times, and the average results are used to determine suitable threshold values (\( T_f \) and \( T_r \)) by minimizing a cost function using a heuristic. Finally, the model is trained on the full dataset and evaluated on a separate test set.}
\label{fig:framework5}
\end{figure*}

To allow a better comprehension of the methodology adopted in this study, Figure \ref{fig:framework5} presents the framework developed. 

Initially, the dataset is split into training and validation sets at a 70:30 ratio to evaluate the proposed framework. The validation set is used to determine a suitable combination of \(T_f\) and \(T_r\), which is later evaluated on external data. Using external test sets aligns with the TRIPOD (\textit{Transparent Reporting of a Multivariable Prediction Model for Individual Prognosis or Diagnosis}) principles, which emphasize the importance of externally evaluating models in healthcare \cite{collins2015transparent}.  

\subsubsection{Extracting instance hardness values }

Following Figure \ref{fig:framework5}, the Instance Hardness (IH) values are calculated for the training data using the Pyhard library \cite{paiva2022relating, paiva2021pyhard}. \textcolor{black}{The algorithms adopted are: Gradient Boosting, Random Forest, Logistic Regression, Multilayer Perceptron, Bagging, and Support Vector Classifier with linear and RBF kernels. The probability of an incorrect classification is obtained via Platt scaling-calibrated log loss \cite{paiva2022relating}.} 

\textcolor{black}{To compute the hardness level of an instance $\mathbf x$, we must compute the probability that $\mathbf x$ is misclassified when the model is trained on the remaining data. Since this type of leave-one-out procedure may be computationally prohibitive, a five-fold cross-validation strategy is adopted. } 
\textcolor{black}{The five-fold cross-validation is repeated five times with different random seeds. 
To reduce computational costs, we did not perform hyperparameter optimization, although this step is possible using the same library. }In the case of missing values, they were imputed using the 3-nearest neighbors. 

Class imbalance is addressed through an undersampling strategy within each cross-validation loop, where instances of the majority class are randomly removed until both classes have the same number of instances. Although the global imbalanced ratio between classes is not a problem, class imbalance exacerbates the effects of other data problems, such as noise and class overlap \cite{batista2004study}.  

\subsubsection{Performing filtering and training the model}

The filtering step is designed to improve data quality by reducing noise and improving class separability. Instances are removed proportionally to the size of each class to avoid creating or worsening class imbalance. Initially, hard instances are progressively removed in increments of 10\%, with up to 50\% of the hardest instances per class. 

Each \(T_f\) applied originates a new dataset used to fit an ML model. The XGBoost algorithm \cite{chen2016xgboost}, chosen for its robust performance, is used throughout the experiments with fixed hyperparameters (see Appendix \ref{apd} for details). 
Undersampling is applied in this stage only when \(n_{\text{min}} / n_{\text{max}} \leq 0.6\), where $n_{\text{min}}$ is the size of the minority class and $n_{\text{max}}$ is the size of the majority class. 

\subsubsection{Rejecting instances}

After filtering and training the model, we test it on the validation set and estimate its confidence. Prediction confidence is derived from the maximum class-conditional probability \(P(Y|X)\) \cite{hendrickx2024machine}. 
To improve the reliability of confidence scores, a sigmoid calibration function is applied using a five-fold cross-validation on the training data. This step ensures that probabilities are well-calibrated while preserving their ranking, enabling more accurate confidence-based decisions \cite{boken2021appropriateness}.

After training and confidence estimation, a range of rejection thresholds (\(T_r\)) from 0.5 to 1.0 (with 1.0 representing full confidence) was tested on the validation set. Increments of 0.02 were used, allowing for a detailed examination of the trade-offs between prediction confidence and coverage. 
The entire process is repeated for five random splits of the dataset into training and validation to ensure robust results. The average results are then used to select the combination of \(T_f\) and \(T_r\) that minimizes the cost function. 

\subsubsection{Selecting the filter and rejection thresholds} \label{sec_heuristic}

The cost function is applied to compare all threshold combinations. It considers three aspects: the model's average performance under a chosen metric, the average confidence, and the rejection rate. For each \(T_f\), we select the \(T_r\) that minimizes the cost.  

All of these are computed as averages over the validation sets. To improve upon this result, a heuristic is applied to explore the neighborhood of the \(T_f\), the value with the lowest observed cost, seeking further improvement. While the heuristic does not guarantee a globally optimal solution, it maintains or improves performance relative to the initial evaluation. 
 
With the final threshold values determined, the model is trained using both the training and validation data, and then evaluated on an external test set.

\subsubsection{Testing configurations}

To assess the effectiveness of our framework, we compare the criteria adopted in both steps — IH for filtering and confidence for rejection — with an alternative measure. Both methods are detailed in the Appendix \ref{apd:baselines}.

To establish a baseline for Instance Hardness values, we assess the extent to which individual training instances negatively impact model performance. An influence-based metric (IF) was adopted, inspired by the framework introduced by Koh and Liang \cite{Koh2017}. This formulation measures how a small up-weighting of an instance affects the validation loss. We begin by computing influence values using a logistic regression model as the base learner. While IH is model-agnostic, influence scores are inherently model-dependent. In this way, we can evaluate how each approach impacts the filtering stage of our strategy. 

For the rejection step, uncertainty is estimated using an ensemble-based approach. This measure is inspired by the approach proposed by \citet{lakshminarayanan2017simple}, which relies on deep ensembles — ensembles of neural networks. Instead of adopting this architecture, a collection of XGBoost models was used. The uncertainty estimation step involves additional computation not required when using the confidence of the main model to reject instances. This allows us to assess whether incorporating a more sophisticated uncertainty measure yields improvements compared to using only model confidence.

In addition to the alternative criteria, we also compare our strategy with the absence of thresholds. In this way, we have four possible setups: 

\begin{enumerate}
    \item \textit{Filtering threshold only:} A dataset, filtered with threshold \(T_f\), is adopted to train the model. During testing, no rejection is performed, all predictions are accepted regardless of their confidence or uncertainty level.
    \item \textit{Rejection threshold only:} No filtering is applied, and the model is trained utilizing all available instances. At test time, a rejection threshold \(T_r\) is applied, filtering out predictions with low confidence or high uncertainty.
    \item \textit{Filtering and Rejecting thresholds:} Using the heuristic, both filtering \(T_f\) and rejection \(T_r\) thresholds are chosen, and the model is trained with a refined version of the dataset. Rejection is adopted, aiming to avoid misclassifications.
    \item \textit{Without thresholds:} The model is trained using all available data without applying any filtering threshold \(T_f\). At inference time, all test instances are considered, with no rejection applied.
\end{enumerate}

These results are presented for each strategy combination. IH and IF are used for filtering, and confidence (C) or uncertainty (U) are adopted for rejection. The threshold values, \(T_f\) and \(T_r\), are determined independently for each configuration through separate evaluations of the cost function. All possible combinations yield 13 configurations.

It is important to note that a direct comparison between these approaches is not straightforward. As performance improvements often come at the cost of an increased rejection rate, this trade-off must be considered when analyzing results. 

\section{Case studies} \label{sec:cases}

This paper presents a framework that can be adapted to meet the user's specific intentions. To demonstrate the full potential of this proposal, we explore it in three case studies that represent different scenarios. They share the characteristic of being routinely collected health-related data, meaning they are observational and not intended for research purposes.

The datasets were assembled with the guidance of a physician, who advised on preprocessing decisions and data interpretation. They are organized into three training–test pairs. The training datasets are used for threshold selection, while the test sets are employed to evaluate the framework's viability on external data.

The origin and pre-processing of each dataset are described in the corresponding subsections, along with the results achieved. 
Table \ref{tab:summary_data} provides an overview of the datasets generated in this work and utilized in our analysis, including the number of instances, features, and the proportion of positive cases.

\begin{table*}[hbtp]
    \centering
    \footnotesize
    \caption{\small Summary of the characteristics of the datasets built in this work, including name of the dataset, number of instances and features, percentage of missing values, and percentage of instances in the positive class. Observ.: Number of observations; Feat.: Number of features; \% NA: percentage of missing values.}
    \begin{tabular}{lcccr}
    \toprule
        \bf Dataset & \bf $\sharp$ Observ. & \bf $\sharp$ Feat. & \bf \% NA & \bf  \bf \% Positive class\\
        \midrule

        \textit{hospitalization\_dengue\_23} & 29,920  & \multirow{2}{*}{37}  & &hospitalized (47.4)   \\
        \textit{hospitalization\_dengue\_24} &  14,345   &  & &hospitalized (39.7) \\ 
        \midrule

        \textit{authorization\_sus} &  19,355  & \multirow{2}{*}{1,047}  & & \multirow{2}{*}{authorized (56.7)}  \\
        \textit{authorization\_sus\_test} &  2,151   & & &  \\

                \midrule

        \textit{severity\_hsl} &  1,432  & \multirow{2}{*}{20} & 6.62 & severe (36.7)   \\
        \textit{severity\_hbp} &  185  & & 8.97 & severe (36.2)\\ 

        \botrule
    \end{tabular}
    
    \label{tab:summary_data}
\end{table*}

We begin our experiments by setting the weights as follows: \( w_p = 4 \), \( w_r = 1 \), and \( w_c = 1 \), yielding an interesting combination that balances performance, rejection rate, and confidence. 

\subsection{Case study 1: hospitalization of patients with Dengue}

The Brazilian Information System for Notifiable Diseases (Sinan) primarily relies on reporting cases of diseases and conditions listed in the national list of notifiable diseases. After gathering information from all health units, the data is made available in batches by year and can be accessed on the Information Technology Department of the Brazilian Unified Health System (DATASUS) website.

One such disease is dengue, which is caused by the Aedes mosquito. Our primary database was from 2023, while data from the beginning of 2024 was adopted as a test set to assess model generalization. 2024 was a year marked by a nationwide increase in disease cases \cite{barcellos2024climate}. The data source contains 1,517,551 notification records from 2023. Notifications are made using a standardized form, resulting in raw data with 121 columns. While some fields are mandatory when reporting a disease occurrence, others may be left incomplete, leading to many blank fields. The dengue data source includes information on the following patient aspects:

\begin{enumerate}
\item Personal details such as gender, professional occupation, race, and education;
\item Dates such as birth date, onset of symptoms, notification of the disease, disease progression, hospitalization, and death;
\item Symptoms, including general symptoms, warning signals, and severe signals;
\item Comorbidities;
\item Location details such as state and health unit.
\end{enumerate}

We have meticulously compiled datasets containing information on dengue across the national territory. This dataset was assembled to predict the need for hospitalization based on the initial symptoms declared in the notification form. The two labels in the dataset are \textit{hospitalized} and \textit{non-hospitalized}. 
An extensive description of the filters applied, data statistics, and final datasets can be found in a GitHub repository \footnote{https://github.com/gabivaleriano/HealthDataBR}. 

The final training dataset contains 29,920 cases and 37 features. \textit{Hospitalized} patients comprise 47.1\% of the cases and 52.9\% are \textit{non-hospitalized} cases. Using data from the first weeks of the following year (2024), we created a test set to evaluate the generalization of models trained on the main dataset. The \textit{hospitalization\_dengue\_24} dataset was assembled following the same preprocessing steps as the main dataset. This test set contains 14,345 cases, of which 39.7\% are infected patients requiring hospitalization, while 60.3\% did not require hospitalization. 

We envisioned an outbreak scenario that would lead to hospital overcrowding. In this context, we proposed a model capable of accurately identifying individuals who will require hospitalization. 
Our approach enables early identification of hospitalization needs shortly after disease notification, facilitating proactive planning and resource allocation. 
The first step was to generate multiple training datasets by filtering out difficult instances in varying proportions and performing a grid search over the filtering thresholds. 

Next, we evaluated model performance using different rejection thresholds. The combinations are compared through the cost function described in Section \ref{sec:evaluating}. A heuristic is employed to explore the neighborhood of the \(T_f\) value that yields the smallest cost. In our main experiment, when using IH values to filter the dataset and performing confidence-based rejections, the smallest cost is achieved with a \(T_f\) value of 0.100. 
The corresponding \(T_r\) that resulted in the optimal cost is 0.940. 

In this dataset, removing a small sample of instances yielded the best predictive performance across five experiment repetitions when evaluating different splits of train-validation sets.  
This suggests that retaining a greater diversity of instances during training helps the model generalize better if model confidence is considered during testing.
These thresholds were then adopted to filter the complete primary dataset, \textit{hosp\_dengue\_23}, and assess performance on the \textit{hosp\_dengue\_24} dataset. 
In Table \ref{tab:performance_metrics}, we report the results of our main strategy in rows 1-3, where it is compared against adopting filtering and rejecting alone, to assess the individual contributions of each step.

\textcolor{black}{Subsequently, we evaluate the use of baseline criteria either isolated or in combination. Rows 4-6 present results obtained using instance hardness values for filtering and uncertainty for rejecting predictions. In rows 7-9, we apply influence values for filtering and confidence values to perform rejection. Finally, we explore the combination of both baselines: IF values for filtering and uncertainty for rejection. In both uncertainty-based scenarios, minimizing the cost function led to no rejections. In this way, we omit rows 4, 5, 10, and 11, and report the filtered results in rows 6 and 12. The standard result, where neither filter nor rejection was performed, is presented in rows 13 and 14. For easier comparison, we report the complement of uncertainty (1 - U), so higher values correspond to greater certainty. }

\begin{table}[h!]
\caption{\small Results of experiments for the XGBoost algorithm trained on the \textit{hospitalization\_dengue\_23} dataset and tested on the \textit{hospitalization\_dengue\_24} dataset. \textbf{F}: Filtering measure; \textbf{R}: Rejection criterion; $\mathbf{T_f}$: Threshold for filtering instances in the training data; $\mathbf{T_r}$: Threshold for rejecting predictions; \textbf{RCL}: Recall; \textbf{PRC}: Precision; \textbf{ACP}: Ratio of accepted predictions; \textbf{C}: Confidence; $\mathbf{U_c}$: Complement of uncertainty (1-U). Experiments 4, 5, 10, and 11 were omitted because minimizing the cost function resulted in no rejection. The best results per metric are highlighted in bold.} \label{tab:performance_metrics}
\centering
\footnotesize
\begin{tabular}{lccccccccccc}
\toprule
& $\mathbf{F}$ & $\mathbf{R}$ & $\mathbf{T_f}$ & $\mathbf{T_r}$ & $\mathbf{PRC_0}$ & $\mathbf{PRC_1}$ & $\mathbf{RCL_0}$ & $\mathbf{RCL_1}$ & $\mathbf{ACP_0}$ & $\mathbf{ACP_1}$ & $\mathbf{C/U_c}$ \\
\midrule

\textbf{1} & \multirow{3}{*}{IH} & \multirow{3}{*}{C} & -- & 0.940 & \textbf{0.951} & \textbf{0.843} & 0.721 & \textbf{0.976} & 0.250 & 0.583 & 0.955 \\
\textbf{2} & & & 0.100 & 0.940 & 0.909 & 0.714 & 0.732 & 0.901 & 0.797 & 0.901 & \textbf{0.988} \\
\textbf{3} & & & 0.100 & -- & 0.884 & 0.656 & 0.704 & 0.860 & \textbf{1.000} & \textbf{1.000} & 0.955 \\
\midrule

\textbf{6} & IH & U & 0.375 & -- & 0.856 & 0.671 & 0.739 & 0.811 & \textbf{1.000} & \textbf{1.000} & \textbf{0.698} \\
\midrule
\midrule

\textbf{7} & \multirow{3}{*}{IF} & \multirow{3}{*}{C} & -- & 0.820 & 0.919 & 0.744 & {0.755} & 0.914 & 0.703 & 0.832 & 0.927 \\
\textbf{8} & & & 0.050 & 0.820 & 0.917 & 0.741 & \textbf{0.757} & 0.910 & 0.721 & 0.837 & {0.929} \\
\textbf{9} & & & 0.050 & -- & 0.880 & 0.665 & 0.719 & 0.851 & \textbf{1.000} & \textbf{1.000} & 0.871 \\
\midrule

\textbf{12} & IF & U & 0.025 & -- & 0.881 & 0.664 & 0.716 & 0.853 & \textbf{1.000} & \textbf{1.000} & 0.602 \\
\midrule
\midrule

\textbf{13} & \multirow{2}{*}{--} & C & \multirow{2}{*}{--} & \multirow{2}{*}{--} & \multirow{2}{*}{0.882} & \multirow{2}{*}{0.664} & \multirow{2}{*}{0.716} & \multirow{2}{*}{0.855} & \multirow{2}{*}{\textbf{1.000}} & \multirow{2}{*}{\textbf{1.000}} & 0.867 \\
\textbf{14} & & U &  & & & & & & & & 0.602 \\
\botrule
\end{tabular}
\end{table}

\textcolor{black}{Table \ref{tab:performance_metrics} presents the results of experiments conducted using the XGBoost algorithm, trained on the \textit{hospitalization\_dengue\_23} dataset and tested on the \textit{hospitalization\_dengue\_24} dataset. In the standard configuration (rows 13–14), the model is trained using all available data without applying filtering or rejection. Under this setting, precision is higher for the negative class (0.882) compared to the positive class (0.664), while recall is reasonably strong for both (0.716 and 0.855, respectively). Since no rejection criterion is applied, the acceptance rate is 100\% for both classes, with moderate model confidence (0.867) and a low complement of uncertainty (0.602).}

\textcolor{black}{The main set of experiments (rows 1–3) adopts IH as the filtering strategy and the level of confidence to reject instances. Applying the rejection step alone (row 1) yields the best predictive performance across all scenarios. However, the model rejects a large portion of instances, classifying only 25\% of negative-class instances and 58\% of positive-class instances. By contrast, training models with filtering only (row 3) improves confidence to 0.955, with performance metrics similar to those of the standard configuration (row 13). When both thresholds are applied (row 2), the best trade-off results are achieved, with negative-class precision and positive-class recall both exceeding 0.9. All remaining metrics are above 0.7. The mean confidence reaches 0.988, while 80\% of negative-class and 90\% of positive-class predictions are accepted. This configuration achieves the best overall trade-off between accuracy and acceptance in this Case study.}

When testing uncertainty as the baseline for rejection (rows 4-6), the thresholds that returned the smallest cost are \(T_f = 0.375\) and no rejection. For this reason, we omitted rows 4 and 5. Applying only filtering (row 6) yields performance similar to row 14, with minor variations.  

\textcolor{black}{In the third scenario, rejection is based on confidence again, while filtering now uses influence values (rows 7–9). The selected thresholds were  \(T_f = 0.050\) and \(T_r = 0.820\). Even the rejection step alone (row 7) improves performance. Precision in the positive class and recall for the negative class are above 0.74, for the remaining metrics values are above 0.91. The mean confidence is 0.927, and the acceptance rates are also solid, surpassing 70\% and 83\% for the negative and positive classes, respectively. Adding the filtering step (row 8) results in marginal changes in the performance metrics and slightly better acceptance rates, which are still lower than the main result in row 2. Filtering alone (row 9) produces a similar profile to the standard (row 13), indicating minimal impact without rejection.}

\textcolor{black}{Finally, in the last scenario, both baselines are combined. Influence values are adopted to filter instances, and rejection is based on uncertainty. Minimizing the cost function indicates that the best strategy is to reject nothing. In this way, rows 10 and 11 were omitted. As the threshold selected to filter the dataset is small, the results are nearly identical to those of the main baseline (row 14).}

\subsection{Case study 2: authorizing specialized medical consultations} \label{dados:telessaude}

The Public National Health System (SUS) provides free healthcare to the entire po\-pu\-la\-tion in Brazil. The system's accessibility and the country's large population lead to a
discrepancy between the supply and demand of specialized
physician consultations \cite{gonccalves2017expanding}. When a patient requires specialized care, general practitioners (GPs) submit a request, which is then assessed for approval by a physician regulator, based on clinical protocols that take into account the severity of the case and the availability of human resources. 

Evaluating these requests requires significant time and effort from experienced clinicians, who must review each case. 
Automating part of this process with an ML system capable of authorizing straightforward cases would alleviate the burden on healthcare teams, saving valuable time and effort, and providing an instant analysis.

This database was accessed through a partnership with Telessaúde-BR, a research center affiliated with the Postgraduate Program in Epidemiology at the Faculdade de Medicina da Universidade Federal do Rio Grande do Sul. With the assistance of the Telessaúde-BR team, who possess in-depth knowledge of the data, a large dataset was assembled. 
The classification task involves predicting which patients will be authorized to receive specialized care.

The database contains a description of the clinical condition in Portuguese, along with patient gender, year of birth, the type of specialized care required, the International Classification of Diseases (ICD), and the final decision made by a clinician, which can be either an authorization or a denial. 
Requests may be denied if the clinical condition does not necessitate specialized care and can be managed by the GP, or if insufficient information is provided to justify the request.

The raw data were prepared using a standard natural language preprocessing pipeline. A detailed description is available in our repository. An important remark is that clinical descriptions were transformed into numerical vectors using a pre-trained BERT model for Brazilian Portuguese. The adopted model is a sentence-transformers model, which maps sentences and paragraphs to a 1024-dimensional dense vector space \cite{souza2020bertimbau}.

This dataset contains no temporal information, making a chronological test split impossible. For this reason, we adopted a random split, stratified by class, selecting a tenth of the instances to compose the test set. The final dataset contains 1,047 features and 19,355 instances, 56.7\% being authorized requests and 43.3\% denied requests. The test set contains 2,151 examples. These datasets are not publicly available since they contain sensitive information. 

In this experiment, if a case is incorrectly denied, it can be appealed within the healthcare system, assuming additional information is provided. However, this may delay the start of treatment. Conversely, an incorrect approval could overload the system, indirectly affecting individuals who genuinely require specialized care. These situations illustrate the importance of achieving strong performance metrics. 

Using the training data, we searched for the thresholds that minimized the cost function. \textcolor{black}{In our main strategy, the value found for the filtering threshold is \(T_f = 0.350\) and the rejection threshold is \(T_r = 0.820\). The results of the final experiment with these thresholds are presented in Table \ref{tab:performance_metrics2}. }

\begin{table}[h!]
\caption{\small Experiments results for the XGBoost algorithm trained on the \textit{authorization\_sus} dataset and tested on the \textit{authorization\_sus\_test} dataset. \textbf{F}: Filtering measure; \textbf{R}: Rejection criterion; $\mathbf{T_f}$: Threshold for filtering instances in the training data; $\mathbf{T_r}$: Threshold for rejecting predictions; \textbf{RCL}: Recall; \textbf{PRC}: Precision; \textbf{ACP}: Ratio of accepted predictions; \textbf{C}: Confidence; $\mathbf{U_c}$: Complement of uncertainty (1-U).} \label{tab:performance_metrics2}
\centering
\footnotesize
\begin{tabular}{lccccccccccc}
\toprule
& $\mathbf{F}$ & $\mathbf{R}$ & $\mathbf{T_f}$ & $\mathbf{T_r}$ & $\mathbf{PRC_0}$ & $\mathbf{PRC_1}$ & $\mathbf{RCL_0}$ & $\mathbf{RCL_1}$ & $\mathbf{ACP_0}$ & $\mathbf{ACP_1}$ & $\mathbf{C/U_c}$ \\
\midrule

\textbf{1} & \multirow{3}{*}{IH} & \multirow{3}{*}{C} & -- & 0.820 & 0.872 & 0.864 & 0.818 & 0.906 & 0.313 & 0.304 & 0.868 \\
\textbf{2} & & & 0.350 & 0.820 & 0.666 & 0.749 & 0.676 & 0.741 & 0.921 & 0.918 & \textbf{0.973} \\
\textbf{3} & & & 0.350 & -- & 0.647 & 0.739 & 0.665 & 0.724 & \textbf{1.000} & \textbf{1.000} & 0.949 \\
\midrule

\textbf{4} & \multirow{3}{*}{IH} & \multirow{3}{*}{U} & -- & 0.860 & 0.915 & \textbf{0.955} & \textbf{0.977} & 0.840 & 0.047 & 0.020 & 0.904 \\
\textbf{5} & & & 0.050 & 0.860 & 0.934 & 0.948 & 0.959 & \textbf{0.917} & 0.079 & 0.049 & 0.919 \\
\textbf{6} & & & 0.050 & -- & 0.676 & 0.728 & 0.622 & 0.772 & \textbf{1.000} & \textbf{1.000} & 0.608 \\
\midrule
\midrule

\textbf{7} & \multirow{3}{*}{IF} & \multirow{3}{*}{C} & -- & 0.860 & 0.889 & 0.887 & 0.902 & 0.872 & 0.219 & 0.148 & 0.888 \\
\textbf{8} & & & 0.025 & 0.860 & 0.883 & 0.877 & 0.872 & 0.888 & 0.260 & 0.204 & 0.891 \\
\textbf{9} & & & 0.025 & -- & 0.693 & 0.725 & 
0.605 & 0.796 & \textbf{1.000} & \textbf{1.000} & 0.750 \\
\midrule

\textbf{10} &  \multirow{3}{*}{IF} & \multirow{3}{*}{U} & -- & 0.920 & \textbf{0.944} & 0.000 & 1.000 & 0.000 & 0.018 & 0.001 & \textbf{0.960} \\
\textbf{11} & & & 0.406 & 0.920 & 0.900 & 0.833 & 0.900 & 0.833 & 0.021 & 0.010 & 0.953 \\
\textbf{12} & & & 0.406 & -- & 0.688 & 0.722 & 0.600 & 0.793 & \textbf{1.000} & \textbf{1.000} & 0.600 \\
\midrule
\midrule

\textbf{13} & \multirow{2}{*}{--} & C & \multirow{2}{*}{--} & \multirow{2}{*}{--} & \multirow{2}{*}{0.688} & \multirow{2}{*}{0.718} & \multirow{2}{*}{0.591} & \multirow{2}{*}{0.796} & \multirow{2}{*}{\textbf{1.000}} & \multirow{2}{*}{\textbf{1.000}} & 0.600 \\
\textbf{14} & & U & & & & & & & & & 0.737 \\
\botrule
\end{tabular}
\end{table}

\textcolor{black}{At the bottom of Table \ref{tab:performance_metrics2} (rows 13-14), we report the performance metrics for models trained on all instances and accepting all predictions. The model demonstrates a relatively poor performance, particularly for the negative class, with a recall of 0.591. The confidence level is low (0.600), while the uncertainty is moderate (complement of uncertainty of 0.737).}

In the first configuration (rows 1-3), we use IH values for filtering, and predictions are rejected according to their confidence levels. In this scenario, applying the rejection threshold alone (row 1) resulted in considerable performance, with all metrics exceeding 0.8, but only a small proportion of instances are classified (around 30\%). Using only the filtering threshold (row 3) increases the model's confidence, and also results in performance metrics improvement compared with the standard result (row 13). The best choice in this scenario, given a trade-off between performance and acceptance ratio, is to combine filtering and rejection (row 2). Although some performance metrics show a slight decrease, gains are mainly in recall for the negative class (the worst metric in row 13). Most instances are accepted, with a rejection rate of less than 10\%. 

\textcolor{black}{In the second configuration (rows 4-6), uncertainty-based rejection is performed. The selected thresholds are \(T_f = 0.050\) and \(T_r = 0.860\). Applying only the rejection threshold (row 4) leads to high predictive performance and a substantial reduction in uncertainty compared to the baseline (row 14). Nonetheless, the model accepts fewer than 5\% of the validation set. Filtering alone yields performance very similar to the standard result (row 14), with an improvement in recall for the negative class. The adoption of both steps —filtering and rejection (row 5)— is the best alternative in this setting, offering high precision and recall, as well as low uncertainty. However, the acceptance rate remains below 10\%, limiting its practical applicability.  }

\textcolor{black}{To test an alternative criterion to the filtering step, we use the IF values in combination with confidence-based rejection (rows 7-9). The heuristic selected a very low filtering threshold \(T_f = 0.025\) and a rejection threshold of \(T_r = 0.860\). Applying only the rejection threshold (row 7) yields strong predictive performance (all metrics above 0.85), albeit at the expense of a low acceptance rate, with 22\% of positive-class and 15\% of negative-class instances accepted. Filtering alone (row 9) results in minor performance variations compared to the standard (row 13). Combining both steps, the performance metrics remain high, and the percentage of accepted instances increases to under 30\%.}

\textcolor{black}{Finally, the fourth configuration (rows 10-12) combines IF values to filter, paired with uncertainty-based rejection. The selected thresholds are  \(T_f = 0.406\) and \(T_r = 0.920\).} 
\textcolor{black}{In this last scenario, when using rejection alone (row 10), the model accepted a small sample of instances, fewer than 2\%. All classifications belong to the negative class, resulting in perfect recall and high precision for this class, and zero performance metrics for the positive class. Filtering alone (row 12) yields performance metrics similar to the standard (row 14). Adopting both steps together (row 11) improves performance, above 0.8, and reduces uncertainty (compared with the standard in row 14), but acceptance rates remain low, 2.1\% for the positive class and 1.0\% for the negative class. }

Although our cost function prioritized predictive performance over confidence and rejection rate, the combination of IH-filtering and confidence-rejection in this Case study yielded low rejection rates and modest improvements in predictive performance metrics. We highlight that  
different weights can be adopted in the cost function, depending on user priorities, allowing flexibility in balancing accuracy, confidence, and coverage. However, a lower rejection rate will inevitably lead to a decline in overall predictive performance, as more low-confidence predictions are accepted. Therefore, the trade-off between rejection and accuracy must be carefully evaluated based on the specific application requirements, ensuring that the model meets operational needs while maintaining a satisfactory level of reliability. 

By automatically handling a significant portion of cases with reliable predictions, our approach helps alleviate system overload and streamline decision-making processes. As a result, the model has been successfully deployed in a real-life setting to partially automate the authorization of specialized medical consultations inside the Brazilian Public Health System. The system is currently in beta, being closely monitored and controlled during this phase. This demonstrates the effectiveness of our approach to improve decision-making while ensuring system reliability.

\subsection{Case study 3: identifying low risk COVID patients} \label{dados:hsl}

The \textit{Covid Data Sharing/BR}  \cite{mello2020opening} repository is a publicly accessible digital database containing laboratory test results collected in hospitals in São Paulo (Brazil) during the pandemic. For each patient who underwent a COVID-19 test, whether hospitalized or receiving ambulatory care, all additional blood tests performed were recorded in the database. 

Among the five available hospitals, we selected two that contain data related to patient outcomes: Hospital Sírio Libanês (HSL) and Hospital Beneficência Portuguesa (HBP). These databases were used to create two distinct datasets with identical features, aimed at predicting severe conditions in hospitalized patients based on the blood tests performed on the day of admission. Data collection in both hospitals started in March 2020. It finishes at different times: in HSL, in May 2021, and in HBP, in January 2021.

The raw data showed notable differences in the lab tests collected for each patient, leading to a high rate of missing values. This may stem from differences in patient profiles across centers and from the lack of a standardized protocol for COVID-19 care and treatment. Additionally, there were variations in test nomenclature and  measurement units which required additional preprocessing strategies. 

Each hospital's raw database consisted of three files containing: 

\begin{itemize}
    \item Demographic patient information, such as birth year and gender.
    \item Laboratory test results performed at the hospital.
    \item Information about the patient's hospital stay and outcomes.
\end{itemize}

The preprocessing steps were designed to integrate, clean, filter, and structure the raw data, ensuring information quality and relevance for predicting COVID-19 severity outcomes.  
In this work, we have adopted the severity criterion as hospitalization lasting more than 14 days or resulting in death. The value was chosen based on the median hospitalization period reported in the literature \cite{Wu2020} and confirmed in the population contemplated in this study. We emphasize that all decisions were made in consultation with a clinical expert. 
The datasets assembled are available in a GitHub repository\footnote{\url{https://github.com/gabivaleriano/HealthDataBR}}, along with documentation describing their construction and statistics.

In a scenario of overcrowded hospitals, identifying upon admission which patients are unlikely to develop severe cases could guide their placement in regular hospital beds rather than intensive care units (ICUs). Rejected patients would undergo further evaluation by a physician and additional tests, while those identified as high risk would receive priority access to more resources. Even partially automating case assessment could reduce hospital burden during peak pandemic periods without compromising patient safety.

The \textit{severity\_hsl} dataset is particularly challenging due to significant class overlap \cite{valeriano2024understanding}. It shows higher standard deviation values among repetitions adopted to select the thresholds. Given its smaller size, it is more sensitive to data-split variability. Figure \ref{fig:comparison} illustrates the variation in the macro-F1 and rejection rate as a function of the rejecting threshold \(T_r\). 

In the main experiment, using instance hardness in the filtering step and performing rejection based on confidence, the minimum cost was achieved with thresholds \(T_f\) = 0.306 and \(T_r\) = 0.880.  Table \ref{tab:performance_metrics3} reports the results for this scenario, as well as for the setups adopting baseline criteria for filtering and rejection. 

\begin{table}[htbp!]
\caption{\small Experiments results for the XGBoost algorithm trained on the \textit{severity\_hsl} dataset and tested on the \textit{severity\_hbp} dataset. \textbf{F}: Filtering measure; \textbf{R}: Rejection criterion; $\mathbf{T_f}$: Threshold for filtering instances in the training data; $\mathbf{T_r}$: Threshold for rejecting predictions; \textbf{RCL}: Recall; \textbf{PRC}: Precision; \textbf{ACP}: Ratio of accepted predictions; \textbf{C}: Confidence; $\mathbf{U_c}$: Complement of uncertainty (1-U).} \label{tab:performance_metrics3}
\centering
\footnotesize
\begin{tabular}{lccccccccccc}
\toprule
& $\mathbf{F}$ & $\mathbf{R}$ & $\mathbf{T_f}$ & $\mathbf{T_r}$ & $\mathbf{PRC_0}$ & $\mathbf{PRC_1}$ & $\mathbf{RCL_0}$ & $\mathbf{RCL_1}$ & $\mathbf{ACP_0}$ & $\mathbf{ACP_1}$ & $\mathbf{C/U_c}$ \\
\midrule

\textbf{1} & \multirow{3}{*}{IH} & \multirow{3}{*}{C} & -- & 0.880 & 0.000 & 0.000 & 0.000 & 0.000 & 0.000 & 0.000 & 0.000 \\
\textbf{2} & & & 0.306 & 0.880 & 0.814 & 0.625 & 0.767 & 0.690 & 0.873 & 0.866 & \textbf{0.966} \\
\textbf{3} & & & 0.306 & -- & 0.777 & 0.575 & 0.737 & 0.627 & \textbf{1.000} & \textbf{1.000} & 0.936 \\
\midrule

\textbf{4} & \multirow{3}{*}{IH} & \multirow{3}{*}{U} & -- & 0.760 & \textbf{1.000} & \textbf{1.000} & \textbf{1.000} & \textbf{1.000} & 0.220 & 0.179 & 0.858 \\
\textbf{5} & & & 0.050 & 0.760 & 0.865 & 0.789 & 0.889 & 0.750 & 0.305 & 0.299 & \textbf{0.872} \\
\textbf{6} &  & & 0.050 & -- & 0.768 & 0.562 & 0.729 & 0.612 & \textbf{1.000} & \textbf{1.000} & 0.672 \\
\midrule
\midrule

\textbf{7} & \multirow{3}{*}{IF} & \multirow{3}{*}{C} & -- & 0.580 & 0.775 & 0.594 & 0.752 & 0.623 & 0.890 & 0.910 & 0.732 \\
\textbf{8} & & & 0.450 & 0.580 & 0.741 & 0.705 & 0.885 & 0.470 & 0.958 & 0.985 & 0.807 \\
\textbf{9} &  & & 0.450 & -- & 0.746 & 0.681 & 0.873 & 0.478 & \textbf{1.000} & \textbf{1.000} & 0.798 \\
\midrule

\textbf{10} & \multirow{3}{*}{IF} & \multirow{3}{*}{U} & -- & 0.660 & 0.854 & 0.833 & 0.911 & 0.741 & 0.381 & 0.403 & 0.790 \\
\textbf{11} & & & 0.050 & 0.660 & 0.902 & 0.833 & 0.920 & 0.800 & 0.424 & 0.373 & 0.812 \\
\textbf{12} & & & 0.050 & -- & 0.772 & 0.577 & 0.746 & 0.612 & \textbf{1.000} & \textbf{1.000} & 0.662 \\
\midrule
\midrule

\textbf{13} & \multirow{2}{*}{--} & C & \multirow{2}{*}{--} & \multirow{2}{*}{--} & \multirow{2}{*}{0.784} & \multirow{2}{*}{0.581} & \multirow{2}{*}{0.737} & \multirow{2}{*}{0.642} & \multirow{2}{*}{\textbf{1.000}} & \multirow{2}{*}{\textbf{1.000}} & 0.712 \\
\textbf{14} & & U &  &  &  &  &  &  &  &  & 0.647 \\
\botrule
\end{tabular}
\end{table}

\textcolor{black}{In Table \ref{tab:performance_metrics3} we present the results of experiments evaluating combinations of filtering training instances and rejecting predictions with an XGBoost model. The model was trained on the \textit{severity\_hsl} dataset and tested on the \textit{severity\_hbp} dataset. The standard results (rows 13 and 14) correspond to the model without any thresholding strategies, exhibiting relatively low precision for the positive class (below 0.6) and only moderate recall (around 0.7).}

\textcolor{black}{In the first set of experiments (rows 1–3), we used our main metrics: IH to filter difficult training instances and a confidence-based rejection during the inference process. The rejection-only strategy (row 1) with a high threshold of \(T_r\) = 0.880 resulted in no predictions being accepted, rendering evaluation impossible. When applying only filtering (row 3), average confidence improves compared to the standard; however, the performance metrics remain suboptimal. Combining filtering and rejection (row 2) led to notable improvements in both precision and recall, particularly for the positive class. This came at the cost of rejecting less than 20\% of the test instances.}

\textcolor{black}{To investigate an alternative to the rejection criterion, we adopted the uncertainty estimates, keeping the IH as the filtering approach (rows 4–6). The thresholds resulting in minimum cost were \(T_f\) = 0.050 and  \(T_r\) = 0.760. }
\textcolor{black}{In this scenario, the rejection strategy alone (row 4) led to a perfect classifier, with fewer than 25\% of test instances accepted. Filtering alone (row 6) has a modest effect on performance and slightly helps to reduce uncertainty when compared with the standard (row 14). The combination of filtering and rejection (row 5) results in strong classification metrics and lower uncertainty, outperforming the confidence-based rejection (row 2). However, the rejection rate exceeds 60\%.}

\textcolor{black}{The third group of experiments (rows 7–9) explores the use of the influence values (IF) as an alternative to the criterion to perform filtering. Rejection is based on confidence again. The thresholds selected in this scenario were \(T_f\) = 0.450 and  \(T_r\) = 0.580.} \textcolor{black}{Rejection alone (row 7), leads to high acceptance rates, however it results in slight changes in most predictive performance metrics, with some increasing and others decreasing, when compared with the standard (row 13). Applying only the filtering step (row 9) results in a decrease in some performance metrics, especially in the recall for the positive class. The selected filtering threshold removed a large number of hard instances from the training dataset, resulting in the loss of important information for model generalization. Combining both strategies (row 8) did not yield a better balance between performance and coverage, as it still presents a low recall for the positive class.}

\textcolor{black}{The final set of results (rows 10–12) combined IF-based filtering with uncertainty rejection. The thresholds \(T_f\) = 0.050 and \(T_r\) = 0.660 yielded the lowest cost. 
Rejection alone (row 10) improved predictive performance metrics and reduced uncertainty (compared to row 14), but fewer than half of all instances were accepted. Filtering alone (row 12) had minimal impact on metrics, yielding only slight changes in either direction. The combined approach (row 11) improved results compared to rejection alone, with more negative-class instances being accepted. Compared with our main configuration (row 2), this method delivered improved predictive performance but at a significant cost in terms of rejection rate.}

\section{Discussion} \label{sec:discussion}

\textcolor{black}{Aggregating the findings across the three case studies reveals some trends that support broader conclusions regarding the interplay between filtering and rejection strategies. This section synthesizes those insights and offers general considerations.
}

\textcolor{black}{Adopting a rejection threshold without prior filtering generally led to the same outcome: there was a considerable improvement in performance metrics and confidence levels, albeit at the cost of accepting fewer instances compared to the combined approach. In contrast, applying only the filtering step typically resulted in modest changes in performance metrics—sometimes improving, sometimes worsening.}

\textcolor{black}{The combined approach, which improved data quality by filtering and rejecting instances at inference time, delivered the best results in most cases.} These overall results are also verified for other benchmark datasets from the health domain in Appendix \ref{apd_results}. The only exception was observed in Case study 3, where using IF values for filtering and confidence for rejection led to a decline in the model's recall for the positive class, possibly due to the over-removal of informative instances. In other scenarios, even minimal dataset refinement, when combined with the rejection step, led to improved performance regardless of the criterion employed. These findings support adopting a two-step framework, as combining filtering and rejection outperforms single-step strategies.

The success of the proposed strategy becomes even more evident in the additional case studies presented in Appendix \ref{apd_results}. We selected benchmark datasets recognized for their high difficulty level \cite{mcelfresh2023neural} and also health-related data. However, they are data collected for research purposes, while our main case studies were derived from routinely collected data. This contrast reinforces the challenging nature of our datasets and highlights the value of our approach in real-world scenarios.

\textcolor{black}{Regarding the choice of criterion for each step, the adoption of uncertainty resulted in no rejection in Case study 1. In Cases 2 and 3, fewer instances were classified when uncertainty was considered, yet the high predictive performance may be also valuable in some scenarios. Confidence scores can be assessed at inference time without additional cost and improve performance metrics in most cases. Given its efficiency and effectiveness, we provide empirical evidence that confidence-based rejection can be an effective practical choice. That said, the framework is flexible, and other rejection criteria can be explored depending on the application. }

Regarding the filtering step, the experiments suggest that adopting IF values can also be a potentially effective approach, achieving performance comparable to or better than our main method, albeit with a higher number of rejected instances. However, results vary across the case studies, and IH-based filtering shows more consistent trends. Since IH employs multiple algorithms, it may improve the stability of the results, though at the cost of higher computational requirements.

We conduct a statistical comparison of the cost values of each solution in Appendix \ref{apd_results}. The cost function was adopted because it is designed to account for the trade-off among performance, rejection rate, and confidence or uncertainty. As the cost is calculated in terms of confidence and uncertainty, we evaluate scenarios separately according to the rejection criterion adopted. While the analysis is limited by the small sample size (N = 6 datasets), the results are consistent with our qualitative analysis, showing that, when considering the cost, IH outperforms the standard solution in both cases. It is also statistically different from IF when combined with confidence-based rejection. 

The influence-based hardness measure introduced in this work shows potential; however, it requires further evaluation and validation. In contrast, the adopted IH formulation has already been deployed across different applications \cite{valeriano2024explaining, valeriano2024understanding, nunes2021using, ferreira2024measuring}, demonstrating robustness and practical reliability. Based on our experiments, it is not easy to definitively favor one approach over the other. At this stage, we suggest the IH approach for practical use, despite its higher computational cost. There is also room for investigating alternative hardness formulations. Additionally, we only adopted XGBoost as the main algorithm in our framework. This choice may affect results, as different algorithms can interact differently with the filtering strategies, potentially impacting both performance and stability.

Lastly, the ethical implications of rejection-based systems are worth highlighting. By allowing models to abstain from non-confident predictions, reject-options contribute to more ethical AI systems that prioritize accuracy and fairness. This is particularly important in applications involving sensitive data, where incorrect predictions can lead to biased outcomes and exacerbate existing inequalities. Selective prediction has been shown to contribute to the development of responsible and ethical AI \cite{kamiran2018exploiting}.

\subsection{Limitations and future directions} 

\textcolor{black}{Despite the promising results presented, our approach has several limitations. As a multi-stage framework, it opens up a vast space of possible configurations and comparisons. In this work, we made deliberate choices to pursue the directions we considered most promising, consequently narrowing the scope of the study. Nevertheless, we believe the potential of our idea can be further explored, and we outline insightful directions for future investigation here. 
}

\textcolor{black}{\textbf{Scope of Experiments and Generalization:} We presented our approach through case studies to emphasize its data-centric nature and adaptability across diverse contexts.} The Appendix \ref{apd_results} enriches our work with three more analyses and statistical tests that align with our main discussion. However, we believe our framework could benefit from a more extensive investigation and a broader empirical evaluation across multiple datasets and varied learners, as we used XGBoost with predefined hyperparameters. A multi-scenario assessment would strengthen our claims and reveal additional performance trends. This investigation may also provide further insights into selecting the most suitable filtering metric. 

\textcolor{black}{\textbf{Extension to multi-class and regression problems:} Our current evaluation focused on binary classification, as we judged the most straightforward scenario to validate initial ideas. However, the approach can be extended to multi-class problems with minimal adjustments. The instance hardness values can be computed in this scenario using the Pyhard library \cite{paiva2022relating}, and confidence-based rejection still holds. Moving to regression problems, it is also possible to estimate hardness values \cite{torquette2022characterizing}. Moreover, in such cases, uncertainty-based rejection may be more suitable than confidence-based methods \cite{sluijterman2024evaluate}.}

\textcolor{black}{\textbf{Ambiguity vs. Novelty Rejection:} Our rejection strategy primarily targets ambiguity, where an input fits both classes and the model is unsure which one is correct.} This is typically reflected by low confidence scores. Another valuable direction is to explore novelty rejection, which addresses inputs that do not resemble any known class or unseen patterns at inference time \cite{hendrickx2024machine}. Uncertainty estimation is capable of supporting novelty when performing rejection \cite{xia2011accurate}. However, there is still room for an in-depth investigation into novelty rejection within the framework \cite{franc2024scod}, particularly concerning a strategy capable of identifying both situations.  

\textcolor{black}{\textbf{Rejection strategies beyond confidence:} Within our two-step framework, any rejection approach can be applied after filtering. We proposed the confidence-rejection for its simplicity and low computational cost. Our results showed that combining filtering and rejection 
improves predictive performance while providing a better balance between performance and a low-rejection rate. In this way, we believe any state-of-the-art reject option classifiers may be improved by adding a filtering step \cite{franc2023optimal}. A promising future direction is to explore an ensemble of learning algorithms for rejection, applying an analogous idea of instance hardness estimation. This would enable rejection from a diverse model perspective, avoid reliance on a single classifier, and foster a data-centered strategy. }

\textcolor{black}{\textbf{Cost function optimization:} The key direction we envision for future research is a deep investigation and better minimization of the cost function. As a first exploration, cost weights were fixed, but other settings could yield substantially different outcomes. Cost weights could be treated as hyperparameters, allowing the cost function to be minimized more effectively. Adding constraints, such as a maximum rejection rate or minimum performance and confidence, can reduce the search space and help direct the search process. }

\textbf{Dynamic filtering via iterative IH computation:} Instance hardness was computed once due to computational constraints; recalculating it iteratively could enable more adaptive filtering. Future work could explore an iterative instance removal strategy, recalculating IH at each step to progressively refine the dataset. Additionally, hyperparameter tuning was not applied to the IH estimation step, meaning our results may not fully exploit the potential of instance filtering. Future studies could adopt this optimization step, already implemented in the Pyhard library \cite{paiva2022relating}, to further enhance generalization. Iterative filtering can also be performed using IF values. 

\section{Conclusion} \label{sec:conclusion}

This study presented and validated a novel data-centric framework for enhancing the reliability of machine learning models, particularly in safety-critical domains such as healthcare. Our method combines instance hardness filtering during training with a confidence-based rejection mechanism at inference time.  \textcolor{black}{As baseline criteria for filtering and rejection, we adopted a hardness measure based on influence values, and an estimation of uncertainty to perform rejection. }By systematically removing hard-to-classify instances and rejecting low-confidence or high-uncertain predictions, our approach ensures that the final model makes more reliable decisions.

The proposed framework enables a trade-off between accuracy and coverage, making it adaptable to different application needs. The key findings from our study include:

\begin{itemize}
    \item Filtering hard instances results in small variations in model performance.
    \item Rejecting low-confident predictions enhances model reliability, although rejecting more instances than in combination with filtering.
    \item Combining filtering and rejection yields a better trade-off between performance and coverage than applying either strategy alone.
    \item \textcolor{black}{Confidence-based rejection presented the better balance between performance and rejection rate.}
    \item \textcolor{black}{Experiments adopting IF values to perform filtering also offered strong results, showing more variability among the studies.}
\end{itemize}

Despite limitations, our work presents a practical and interpretable framework that balances data quality enhancement and prediction confidence. This approach is particularly relevant for medical AI applications, where rejecting a low-confidence decision is preferable to making an incorrect one. By enabling ML models to identify and abstain from making predictions, our framework supports safer and more trustworthy AI deployment in real-world scenarios.

\section*{Declarations}

\subsection*{Funding}
This work was partially supported by the Brazilian research agencies Coordenação de Aperfeiçoamento de Pessoal de Nível Superior – Brasil (CAPES) – under the grant "Edital CAPES 12/2020 Telemedicina e Análise de Dados Médicos (88881.507039/2020-01,
88887.507037/2020-00)" and Fundação de Amparo à Pesquisa do Estado de São Paulo (FAPESP) Process Number \#2021/06870-3.

\subsection*{Conflicts of interest}
The authors declare that they have no conflict of interest.

\subsection*{Availability of data and material}
Four datasets adopted in this work (\textit{severity\_hsl}, \textit{severity\_hbp}, \textit{hospitalization\_dengue\_23} and \textit{hospitalization\_dengue\_24}) are available in a public repository on Github, along with their corresponding documentation (\url{https://github.com/gabivaleriano/HealthDataBR}). The datasets from Case Study 2 are not available as they may contain sensitive information. The datasets adopted in Appendix \ref{apd_results} are publicly available in the OpenML repository. For reproducibility, we replicated them in our repository using our specific train–test splits.

\subsection*{Code availability}
The code for the experiments and the corresponding results are available in a public GitHub repository (\url{https://github.com/gabivaleriano/RefineAI}).

\subsection*{Ethics approval}
The data sources for Case Studies 1 and 3 are anonymized. The dataset for Case Study 2 was approved by the ethics committee under process number 02658318.8.0000.5327. 

\subsection*{Consent to participate}
Not Applicable.

\subsection*{Consent for publication}
Not Applicable.

\subsection*{Authors' contributions}
M.G.V. assembled the datasets used in the experiments, conceived the main idea of the paper, implemented the code for running the experiments, and drafted the manuscript. D.K.M. contributed to the original concept of the paper and supervised part of the work. A.M. and N.K. provided expertise in the analysis of Case Study 2. C.R.V.K. guided the data preparation process and assisted in the analysis of the results. A.C.L. supervised the entire project and reviewed the paper. All authors contributed to the writing and organization of the paper.

\input sn-article.bbl

\begin{appendices}

\section{Additional Case studies} \label{apd_results}

To complement the main experiments, we present three additional Case studies in this appendix. The datasets were selected from a list provided by Tabzilla, a benchmark of hard-to-classify tabular datasets presented in McElfresh et al. \cite{mcelfresh2023neural}. In this benchmark, the criterion for considering a dataset difficult is the predictive performance of different ensemble algorithms. From a collection of experiments using the OpenML repository \cite{vanschoren2014openml}, they report a list of 36 challenging datasets. We selected three datasets from the health area to ensure consistency in our analysis. 

To replicate our methodology, categorical features were one-hot encoded, as the algorithms used to calculate IH and IF cannot handle categorical variables directly. The datasets were split into training (70\%) and test (30\%) sets, and all preprocessing scripts and data partitions are available in our repository for reproducibility. Table \ref{tab:summary_data_apd} presents a summary of the datasets' statistics. Notably, there are no missing values.

\begin{table*}[hbtp]
    \centering
    \footnotesize
    \caption{\small Summary of the characteristics of the datasets that compose the additional Case studies, including the name of the dataset, the number of instances and features, the percentage of missing values, and instances in the positive class. Observ.: Number of observations; Feat.: Number of features.}
    \begin{tabular}{lccr}
    \toprule
        \bf Dataset & \bf $\sharp$ Observ. & \bf $\sharp$ Feat. &  \bf  \bf \% Positive class\\
        \midrule

        \textit{lymph\_train} &  102  & \multirow{2}{*}{35} &  \multirow{2}{*}{normal (58.8)}\\
        \textit{lymph\_test} &  46  &   & \\ 
        \midrule

        \textit{heart\_train} & 205  & \multirow{2}{*}{23}  & \multirow{2}{*}{disease (34.0)}  \\
        \textit{heart\_test} &  89   &   & \\ 
        \midrule

        \textit{bioresponse\_train} &  2,624  & \multirow{2}{*}{1,777}  & \multirow{2}{*}{bioresponse (54.2)} \\
        \textit{bioresponse\_test} &  1,127   &  &   \\ 
        \botrule
    \end{tabular}
    
    \label{tab:summary_data_apd}
\end{table*}

Each classification task and the experimental results are described next.

\subsection{Normal lymph node identification}

The Lymphography dataset, collected at the University Medical Centre Ljubljana and contributed by Kononenko and Cestnik in 1988, has been widely used in the ML literature \cite{cestnik1987, clark1987, michalski1986}. The dataset consists of 148 instances with 19 attributes (including the target feature) that describe the structural and pathological properties of lymph nodes observed in medical imaging. 

The classification task involves four categories: \textit{normal}, \textit{metastases}, \textit{malignant lymph}, and \textit{fibrosis}. To enable binary classification, the target classes were reduced to \textit{normal} and \textit{abnormal}. After applying our complete framework, the filtering and rejecting thresholds selected in our main strategy were: \(T_f\) = 0.100 and \(T_r\) = 0.900. The results of applying these thresholds to the training dataset and evaluating them on the test set are reported in Table \ref{tab:performance_lymph}, which also presents the baseline results for each step.

\begin{table}[h!]
\caption{\small Experiments results for the XGBoost algorithm trained on the \textit{lymph\_train} dataset and tested on the \textit{lymph\_test} dataset. \textbf{F}: Filtering measure; \textbf{R}: Rejection criterion; $\mathbf{T_f}$: Threshold for filtering instances in the training data; $\mathbf{T_r}$: Threshold for rejecting predictions; \textbf{RCL}: Recall; \textbf{PRC}: Precision; \textbf{ACP}: Ratio of accepted predictions; \textbf{C}: Confidence; $\mathbf{U_c}$: Complement of uncertainty (1-U). Experiments 11 and 12 were omitted since the cost optimization returned no filtering as the best strategy for this scenario.} \label{tab:performance_lymph}
\centering
\footnotesize
\begin{tabular}{lccccccccccc}
\toprule
& $\mathbf{F}$ & $\mathbf{R}$ & $\mathbf{T_f}$ & $\mathbf{T_r}$ & $\mathbf{PRC_0}$ & $\mathbf{PRC_1}$ & $\mathbf{RCL_0}$ & $\mathbf{RCL_1}$ & $\mathbf{ACP_0}$ & $\mathbf{ACP_1}$ & $\mathbf{C/U_c}$ \\
\midrule

\textbf{1} & \multirow{3}{*}{IH} & \multirow{3}{*}{C} & -- & 0.900 & \textbf{1.000} & \textbf{1.000} & \textbf{1.000} & \textbf{1.000} & 0.421 & 0.148 & 0.909 \\
\textbf{2} & & & 0.100 & 0.900 & \textbf{1.000} & 0.933 & 0.933 & \textbf{1.000} & 0.789 & 0.519 & \textbf{0.942} \\
\textbf{3} & & & 0.100 & -- & 0.708 & 0.909 & 0.895 & 0.741 & \textbf{1.000} & \textbf{1.000} & 0.873 \\

\midrule
\textbf{4} & \multirow{3}{*}{IH} & \multirow{3}{*}{U} & -- & 0.660 & 0.786 & 0.923 & 0.917 & 0.800 & 0.632 & 0.556 & 0.835 \\
\textbf{5} & & & 0.100 & 0.660 & 0.812 & 0.923 & 0.929 & 0.800 & 0.737 & 0.556 & \textbf{0.843} \\
\textbf{6} & & & 0.100 & -- & 0.708 & 0.909 & 0.895 & 0.741 & \textbf{1.000} & \textbf{1.000} & 0.739 \\

\midrule
\midrule
\textbf{7} & \multirow{3}{*}{IF} & \multirow{3}{*}{C} & -- & 0.860 & 0.867 & \textbf{1.000} & \textbf{1.000} & 0.833 & 0.684 & 0.444 & 0.896 \\
\textbf{8} & & & 0.050 & 0.860 & 0.667 & 0.929 & 0.933 & 0.650 & 0.789 & 0.741 & {0.911} \\
\textbf{9} & & & 0.050 & -- & 0.600 & 0.938 & 0.947 & 0.556 & \textbf{1.000} & \textbf{1.000} & 0.865 \\

\midrule


\textbf{10} & IF & U & -- & 0.560 & 0.609 & 0.933 & 0.933 & 0.609 & 0.789 & 0.852 & {0.766} \\
\midrule
\midrule

\textbf{13} & \multirow{2}{*}{--} & C & \multirow{2}{*}{--} & \multirow{2}{*}{--} & \multirow{2}{*}{0.600} & \multirow{2}{*}{0.938}  & \multirow{2}{*}{0.947} & \multirow{2}{*}{0.556} & \multirow{2}{*}{\textbf{1.000}} & \multirow{2}{*}{\textbf{1.000}} & {0.822} \\
\textbf{14} & & U & & &  &  &  &  &  &  & 0.723 \\
\botrule
\end{tabular}
\end{table}

The model performance on the raw dataset (rows 13 and 14) is imbalanced among classes. The recall for the negative class and precision for the positive class are high, while the remaining metrics are unsatisfactory. The adoption of filtering or rejection alone led to improvements in predictive performance metrics in some scenarios (rows 1, 3, 4, 6, and 7). However, applying them together results in a better balance between performance and rejection rate. 

Across scenarios, our main setup (row 2) presents the best approach. All metrics are above 0.9, and a considerable number of instances are classified, 79\% and 52\% of acceptance for the negative and positive classes, respectively. Notably, the recall of the positive class (the worst metric in row 13) significantly improved with the proposed strategy. The combination of IH filtering with uncertainty rejection also offered strong performance metrics while accepting a slightly smaller number of instances. Adopting IF to filter and confidence-based rejection increased the percentage of instances accepted in the positive class, but did not offer the same robustness in the metrics.

\subsection{Heart disease prediction}

The Heart Disease (Hungarian) dataset, provided by Andras Janosi, M.D. from the Hungarian Institute of Cardiology, Budapest, is a reprocessed version of the Hungarian subset of the UCI Heart Disease dataset \cite{asuncion2007uci}. The dataset comprises patient-level clinical features and diagnostic test results commonly used to assess heart disease. The prediction task is formulated as a binary classification problem, where the positive class corresponds to patients diagnosed with heart disease.

In our principal strategy, when adopting IH to perform filtering and confidence-based rejection, the selected thresholds are \(T_f\) = 0.006 and \(T_r\) = 0.740. The results of applying these thresholds are summarized in \ref{tab:performance_heart}.

\begin{table}[h!]
\caption{\small Experiments results for the XGBoost algorithm trained on the \textit{heart\_train} dataset and tested on the \textit{heart\_test} dataset. \textbf{F}: Filtering measure; \textbf{R}: Rejection criterion; $\mathbf{T_f}$: Threshold for filtering instances in the training data; $\mathbf{T_r}$: Threshold for rejecting predictions; \textbf{RCL}: Recall; \textbf{PRC}: Precision; \textbf{ACP}: Ratio of accepted predictions; \textbf{C}: Confidence; $\mathbf{U_c}$: Complement of uncertainty (1-U). Experiments 8, 9, 11, and 12 were omitted since the cost optimization returned no filtering as the best strategy for these scenarios.} \label{tab:performance_heart}
\centering
\footnotesize
\begin{tabular}{lccccccccccc}
\toprule
& $\mathbf{F}$ & $\mathbf{R}$ & $\mathbf{T_f}$ & $\mathbf{T_r}$ & $\mathbf{PRC_0}$ & $\mathbf{PRC_1}$ & $\mathbf{RCL_0}$ & $\mathbf{RCL_1}$ & $\mathbf{ACP_0}$ & $\mathbf{ACP_1}$ & $\mathbf{C/U_c}$ \\
\midrule

\textbf{1} & \multirow{3}{*}{IH} & \multirow{3}{*}{C} & -- & 0.740 & 0.500 & 0.986 & 0.750 & 0.958 & 0.800 & 0.845 & 0.910 \\
\textbf{2} & & & 0.006 & 0.740 & 0.600 & 0.975 & 0.600 & 0.975 & \textbf{1.000} & 0.952 & \textbf{0.940} \\
\textbf{3} & & & 0.006 & -- & 0.600 & 0.976 & 0.600 & 0.976 & \textbf{1.000} & \textbf{1.000} & 0.926 \\
\midrule

\textbf{4} & \multirow{3}{*}{IH} & \multirow{3}{*}{U} & -- & 0.580 & \textbf{1.000} & 0.984 & 0.750 & \textbf{1.000} & 0.800 & 0.750 & 0.712 \\
\textbf{5} & & & 0.006 & 0.580 & \textbf{1.000} & 0.986 & 0.750 & \textbf{1.000} & 0.800 & 0.833 & \textbf{0.793} \\
\textbf{6} & & & 0.006 & -- & 0.600 & 0.976 & 0.600 & 0.976 & \textbf{1.000} & \textbf{1.000} & 0.747 \\
\midrule
\midrule

\textbf{7} & IF & C & -- & 0.740 & 0.500 & 0.986 & 0.750 & 0.958 & 0.800 & 0.845 & 0.910 \\
\midrule

\textbf{10} & {IF} & {U} & -- & 0.560 & \textbf{1.000} & 0.985 & 0.750 & \textbf{1.000} & 0.800 & 0.786 & 0.706 \\

\midrule
\midrule

\textbf{13} & \multirow{2}{*}{--} & C & \multirow{2}{*}{--} & \multirow{2}{*}{--} & \multirow{2}{*}{0.250} & \multirow{2}{*}{0.974} & \multirow{2}{*}{0.600} & \multirow{2}{*}{0.893} & \multirow{2}{*}{\textbf{1.000}} &\multirow{2}{*}{\textbf{1.000}}  & 0.864 \\
\textbf{14} & & U& & & & & & & & & 0.668 \\
\botrule
\end{tabular}
\end{table}

In the standard scenario, the model's performance is imbalanced among classes (rows 13 and 14). The precision in the negative class is very low, at 0.250. Applying either filtering or rejecting alone improved this specific metric in most cases (rows 1, 3, 4, 6, 7, and 10). The adoption of uncertainty-based rejection proved extremely useful in this dataset, resulting in high metrics and acceptance ratio, even when adopted alone (rows 4 and 10). Our main strategy (row 2) yielded a high acceptance rate but intermediate predictive performance metrics. Considering both performance and rejection rates, the best strategy appears to be filtering with IH values and performing uncertainty-based rejection (row 5), which resulted in high predictive performance metrics with more than 80\% of acceptance. Adopting IH to filter and rejecting based on uncertainty also offered a strong classifier, while accepting fewer instances (compared to row 2). 

\subsection{Detecting molecules bioresponse}

The Bioresponse dataset is designed to predict biological responses of molecules based on their chemical properties. Each row corresponds to a molecule, with the target attribute indicating whether the molecule elicited a biological response (1) or not (0). The dataset contains 1,776 physicochemical molecular descriptors, calculated to capture structural features such as size, shape, and composition. These descriptors have been normalized to facilitate learning. 

In the main experiment, the thresholds obtained by optimizing the cost function were \(T_f\) = 0.175 and \(T_r\) = 0.860. Table \ref{tab:performance_bio} summarizes the results for individual and combined application of the thresholds, alongside the baseline criteria for filtering and rejecting. 

\begin{table}[h!]
\caption{\small Experiments results for the XGBoost algorithm trained on the \textit{bioresponse\_train} dataset and tested on the \textit{bioresponse\_test} dataset. \textbf{F}: Filtering measure; \textbf{R}: Rejection criterion; $\mathbf{T_f}$: Threshold for filtering instances in the training data; $\mathbf{T_r}$: Threshold for rejecting predictions; \textbf{RCL}: Recall; \textbf{PRC}: Precision; \textbf{ACP}: Ratio of accepted predictions; \textbf{C}: Confidence; $\mathbf{U_c}$: Complement of uncertainty (1-U). Experiments 8 and 9 were omitted since the cost optimization returned no filtering as the best strategy for this scenario.} \label{tab:performance_bio}
\centering
\footnotesize
\begin{tabular}{lccccccccccc}
\toprule
& $\mathbf{F}$ & $\mathbf{R}$ & $\mathbf{T_f}$ & $\mathbf{T_r}$ & $\mathbf{PRC_0}$ & $\mathbf{PRC_1}$ & $\mathbf{RCL_0}$ & $\mathbf{RCL_1}$ & $\mathbf{ACP_0}$ & $\mathbf{ACP_1}$ & $\mathbf{C/U_c}$ \\
\midrule
\textbf{1} & \multirow{3}{*}{IH} & \multirow{3}{*}{C} & -- & 0.860 & \textbf{0.923} & \textbf{0.905} & \textbf{0.840} & \textbf{0.956} & 0.413 & 0.553 & 0.883 \\
\textbf{2} & & & 0.175 & 0.860 & 0.799 & 0.809 & 0.767 & 0.837 & 0.915 & 0.912 & \textbf{0.963} \\
\textbf{3} & & & 0.175 & --   & 0.781 & 0.795 & 0.748 & 0.823 & \textbf{1.000} & \textbf{1.000} & 0.941 \\
\midrule

\textbf{4} & \multirow{3}{*}{IH} & \multirow{3}{*}{U} & -- & 0.540 & 0.822 & 0.823 & 0.770 & 0.866 & 0.828 & 0.864 & 0.748 \\
\textbf{5} & & & 0.387 & 0.540 & 0.782 & 0.776 & 0.726 & 0.824 & 0.948 & 0.920 & \textbf{0.867} \\
\textbf{6} & & & 0.387 & --   & 0.749 & 0.773 & 0.723 & 0.795 & \textbf{1.000} & \textbf{1.000} & 0.843 \\
\midrule
\midrule

\textbf{7} & IF & C & -- & 0.780 & 0.865 & 0.867 & 0.814 & 0.905 & 0.700 & 0.791 & \textbf{0.864} \\
\midrule

\textbf{10} & \multirow{3}{*}{IF} & \multirow{3}{*}{U} & -- & 0.520 & 0.816 & 0.808 & 0.754 & 0.859 & 0.891 & 0.908 & 0.735 \\
\textbf{11} & & & 0.025 & 0.520 & 0.828 & 0.808 & 0.754 & 0.868 & 0.890 & 0.895 & \textbf{0.745} \\
\textbf{12} & & & 0.025 & --   & 0.807 & 0.794 & 0.738 & 0.851 & \textbf{1.000} & \textbf{1.000} & 0.719 \\
\midrule
\midrule

\textbf{13} & \multirow{2}{*}{--} & C & \multirow{2}{*}{--} & \multirow{2}{*}{--} & \multirow{2}{*}{0.802} & \multirow{2}{*}{0.793} & \multirow{2}{*}{0.738} & \multirow{2}{*}{0.846} & \multirow{2}{*}{\textbf{1.000}} & \multirow{2}{*}{\textbf{1.000}} & 0.814 \\
\textbf{14} & & U& & & & & & & & & 0.712 \\
\botrule
\end{tabular}
\end{table}

For this dataset, the standard predictive performance is intermediate, with all metrics over 0.73. Applying only filtering did not result in significant benefits, except for increasing average confidence in row 3. The adoption of rejection alone resulted in improved model performance (rows 1, 4, 7, and 10). From a global perspective, the best result is filtering with IH values and rejection based on confidence (row 2). This strategy consistently improves performance while accepting the majority of instances (fewer than 10\% rejection). The threshold combinations in rows 5 and 11 also yielded strong results, offering a balance between performance and rejection rate that is close to that of our main setup.

\subsection{A statistical comparison among methods}

To compare the methods statistically, we used a Friedman test \cite{demvsar2006statistical}. The objective was to evaluate the standard results against each of the criterion combinations for filtering and rejection across the six case studies. The main challenge of this investigation is balancing improvements in predictive performance with the rate of accepted instances. 

To select the thresholds in each step, we adopted a cost function, described in Section \ref{sec:evaluating}. The cost incorporates three aspects of the framework evaluation: model performance, rejection rate, and confidence or uncertainty. Now, to evaluate the results across different configurations, the cost of each solution serves as a suitable metric to account for the trade-off involved in selective classification. 

This approach is limited, as it is not fair to directly compare confidence and uncertainty values, since they capture different aspects of the data and are not straightforwardly comparable. In this way, we conducted the statistical test in two steps, evaluating confidence and uncertainty methods separately. Despite this limitation, we found the cost function to be the fairest option for this evaluation. 

The Friedman test is a nonparametric statistical test designed to detect differences among multiple algorithms across datasets \cite{demvsar2006statistical}. We treat each criterion combination as a distinct method for applying the test. It first ranks algorithms by performance to determine whether the difference in cost values is genuinely significant or just due to chance. 

The test is used to test evidence for rejecting the null hypothesis, that is: all the methods perform equally. In both comparisons the null-hypothesis is rejected, offering evidence that the results of different methods do not belong to the same distribution, with p-value $< 0.033$ and 0.025 for the uncertainty and confidence methods. Since our sample is small ($N = 6$ datasets), the statistical test provides only weak evidence; however, we found it worth reporting, as the results agree with our previous interpretation and descriptive analysis for each case study. 

\begin{figure*}[htbp!]
    \centering

    \begin{minipage}{0.495\textwidth}
        \centering
        \includegraphics[width=\textwidth]{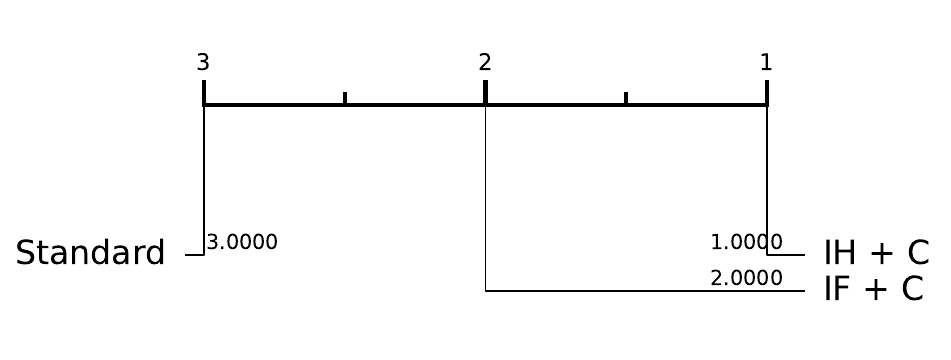}
        \caption*{(a) Confidence methods.}
        \label{fig:image1}
    \end{minipage}
    \hfill
    \begin{minipage}{0.495\textwidth}
        \centering
        \includegraphics[width=\textwidth]{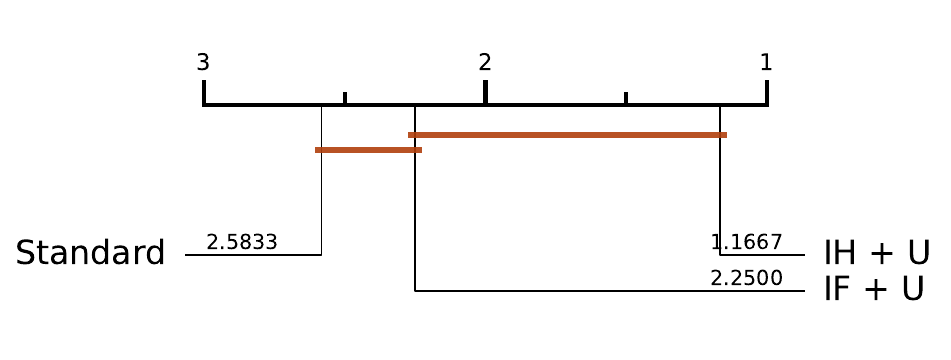}
        \caption*{(b) Uncertainty methods.}
        \label{fig:image2}
    \end{minipage}

    \caption{\small Critical difference diagrams comparing confidence-based methods (left) and uncertainty-based methods (right). In the first image, the three methods are statistically different; in the second diagram only IH+U differ from the standard. } \label{fig:cd}
\end{figure*}

To compare which method is better, we applied a Wilcoxon post hoc test with Holm correction, comparing the combination of filter and rejection with the standard cost values. The critical difference diagrams in Figure \ref{fig:cd} illustrate which methods presented statistically significant results in each comparison. 

In the diagrams, algorithms connected by a horizontal bar are not significantly different from each other based on the statistical test. In Figure \ref{fig:cd}(a), on the left, the confidence-based methods all differ from each other, with the method adopting IH filtering presenting the best performance, followed by IF filtering, and finally the standard case where no filtering or rejection is performed. 

On the left, Figure \ref{fig:cd}(b) shows the results for the uncertainty methods; here, the only statistically different one is the filtering with IH when compared with the standard scenario with no filtering or rejection. Those results reinforce our analysis in Section \ref{sec:discussion}, namely that filtering with IH showed a more consistent trend and that more tests are needed to confirm this superiority over IF filtering.                                 

\section{Baselines} \label{apd:baselines}

This appendix provides a more detailed description of the baselines adopted in our study. 

\subsection{\textcolor{black}{Influence based measure}}\label{sec:IF}

To assess the degree to which individual training instances negatively affect model performance, we employ an influence-based metric (IF) inspired by the framework introduced by Koh and Liang \cite{Koh2017}. Our approach is specifically tailored to identify harmful instances, those that degrade model generalization when included in the training process. 

We begin by computing influence values using a logistic regression model as the base learner. This model is compatible with the theoretical framework, which assumes a differentiable, convex loss function, such as the logistic loss, and a relatively low-dimensional parameter space \cite{Koh2017}. 

Given a trained model with parameters $\hat{\theta}$ and a training instance $z_i = (\mathbf x_i, y_i)$, we approximate the effect of up-weighting $z_i$ on the loss incurred at a validation point $z_{\text{val}} = (\mathbf x_{\text{val}}, y_{\text{val}})$:
\[
\mathcal{I}_{\text{up,loss}}(z_i, z_{\text{val}}) = -\nabla_{\theta} L(z_{\text{val}}, \hat{\theta})^\top H_{\hat{\theta}}^{-1} \nabla_{\theta} L(z_i, \hat{\theta})
\]

where:
\begin{itemize}
    \item $\mathcal{I}_{\text{up,loss}}(z_i, z_{\text{val}})$ is the Influence value of instance $z_i$ on a validation point $z_{\text{val}}$,
    \item $L(z, \theta)$ is the loss function (e.g., logistic loss),
    \item $\nabla_{\theta} L(z, \hat{\theta})$ is the gradient of the loss with respect to the model parameters,
    \item $H_{\hat{\theta}} = \frac{1}{n} \sum_i \nabla^2_{\theta} L(z_i, \hat{\theta})$ is the empirical Hessian of the loss over the training data,
    \item $H_{\hat{\theta}}^{-1}$ is the inverse Hessian, capturing local curvature information.
\end{itemize}

This formulation measures how a small up-weighting of $z_i$ affects the validation loss. To approximate the effect of removing or down-weighting an instance, we flip the sign of the influence values. This enables us to estimate whether a training instance contributes positively or negatively to model generalization.

We apply this analysis in a 5-fold cross-validation setting, repeated five times with different random seeds. In each iteration, one fold is held out for validation, and the model is trained on the remaining data. Influence scores are calculated per fold. 

To reinforce the relationship between harmful influence and poor model performance, each estimated influence value is scaled by the corresponding log-loss of the validation fold. This ensures that harmful instances in the training set that led to higher errors on the validation set are weighted more strongly. This weighting aligns the influence estimation with the severity of model error in that fold. Higher values indicate greater harm to generalization. Values are scaled to the $[0,1]$ interval using min-max normalization.

Scaled influence values are averaged across all 25 cross-validation runs, producing a robust, normalized estimate of each instance’s harmfulness.
Since class imbalance is addressed via random under-sampling inside each loop of the cross-validation, some instances may not be selected in any fold. We handle this by performing an additional influence-estimation step that ensures coverage of all instances.

This influence-based score provides a dataset-contextual measure of how detrimental a training example is, on average, to model generalization. Unlike IH values, which reflect the classification difficulty of an instance averaged among a pool of diverse learners, our influence-based approach measures how much each example contributes to increasing validation loss across different train-validation configurations.
Our goal is to introduce a complementary perspective and assess its benefits for our framework.

\subsection{{\textcolor{black}{Uncertainty estimation}}}\label{sec:U}

For the rejection step, uncertainty is estimated using an ensemble-based approach. This measure is inspired by the approach proposed by \citet{lakshminarayanan2017simple}, which relies on deep ensembles — ensembles of neural networks.

Since our objective was to design a method that also operates effectively on small datasets, such as those used in Case Study 3, we constructed a meta-ensemble of XGBoost models with varied hyperparameters, effectively creating a higher-level ensemble. This strategy was adopted instead of neural networks, which may overfit in low-data regimes and typically require high computational resources.

In our method, a collection of XGBoost models, each consisting of only 15 estimators, is trained with varying hyperparameters. The goal was to construct weak learners and evaluate the degree of divergence in their predictions. Using the same training instances, each model predicts class probabilities on the validation set. Then we compute the entropy of the predicted probabilities across models for each validation instance. 

For instances that are well represented in the training data, far from the decision boundary and surrounded by other instances of the same class, we expect all models to make similar, confident predictions, resulting in low entropy. Conversely, when models disagree on the predicted class or assign diffuse probability distributions, the entropy increases, indicating greater uncertainty. Values are scaled to the $[0.5,1]$ interval using min-max normalization, aligning with confidence scores and facilitating a more direct comparison.

Table \ref{tab:xgb_params} presents the configuration of the ensemble used to assess uncertainty. 

\begin{table}[htbp!]
\centering
\caption{\small List of hyperparameters used for the ensemble of XGBoost models.} 
\label{tab:xgb_params}
\begin{tabular}{cccccc}
\toprule
\textbf{max\_depth} & \textbf{learning\_rate} & \textbf{subsample} & \textbf{colsample\_bytree} & \textbf{random\_state} \\
\midrule
3 & 0.10 & 0.80 & 0.80 & 0 \\
4 & 0.05 & 0.70 & 1.00 & 1 \\
5 & 0.20 & 1.00 & 0.60 & 2 \\
6 & 0.10 & 0.90 & 0.90 & 3 \\
3 & 0.30 & 0.60 & 0.70 & 4 \\
4 & 0.15 & 0.85 & 1.00 & 5 \\
5 & 0.07 & 0.75 & 0.85 & 6 \\
6 & 0.10 & 0.65 & 0.90 & 7 \\
3 & 0.20 & 0.95 & 0.60 & 8 \\
4 & 0.10 & 0.80 & 0.95 & 9 \\
\bottomrule
\end{tabular}
\end{table}

\section{\textcolor{black}{Minimizing the cost function}} \label{apd_cost}

The definition of a cost function enabled the integration of three critical aspects into a single evaluation: confidence, rejection rate, and performance. For this reason, by optimizing costs, we enabled objective selection of the filtering and rejection thresholds. 

To determine a suitable optimization strategy, we conducted an empirical evaluation of the cost function to understand its behavior across different threshold values better. This evaluation was conducted in our smallest dataset to offer insights that, although not conclusive, help justify the adoption of the heuristic proposed in Section \ref{sec_heuristic}. Due to its computational demands, this type of investigation is impractical to apply to the remaining datasets. 

The critical step in selecting thresholds is defining the proportion of instances to filter. Once the dataset is refined and the model is trained, all rejection thresholds can be easily evaluated using the validation set, and the best one is selected. 

\begin{figure*}[htbp]
    \centering

    \begin{minipage}{0.495\textwidth}
        \centering
        \includegraphics[width=\textwidth]{images/cost_vs_threshold_macro-f1_ih_confidence_grid.pdf}
        \caption*{(a) Grid search.}
        \label{fig:image1_apd}
    \end{minipage}
    \hfill
    \begin{minipage}{0.495\textwidth}
        \centering
        \includegraphics[width=\textwidth]{images/cost_vs_threshold_macro-f1_ih_confidence_heuristic.pdf}
        \caption*{(b) Grid search and heuristic}
        \label{fig:image2_apd}
    \end{minipage}

    \vspace{0.4cm} 

    \begin{minipage}{0.495\textwidth}
        \centering
        \includegraphics[width=\textwidth]{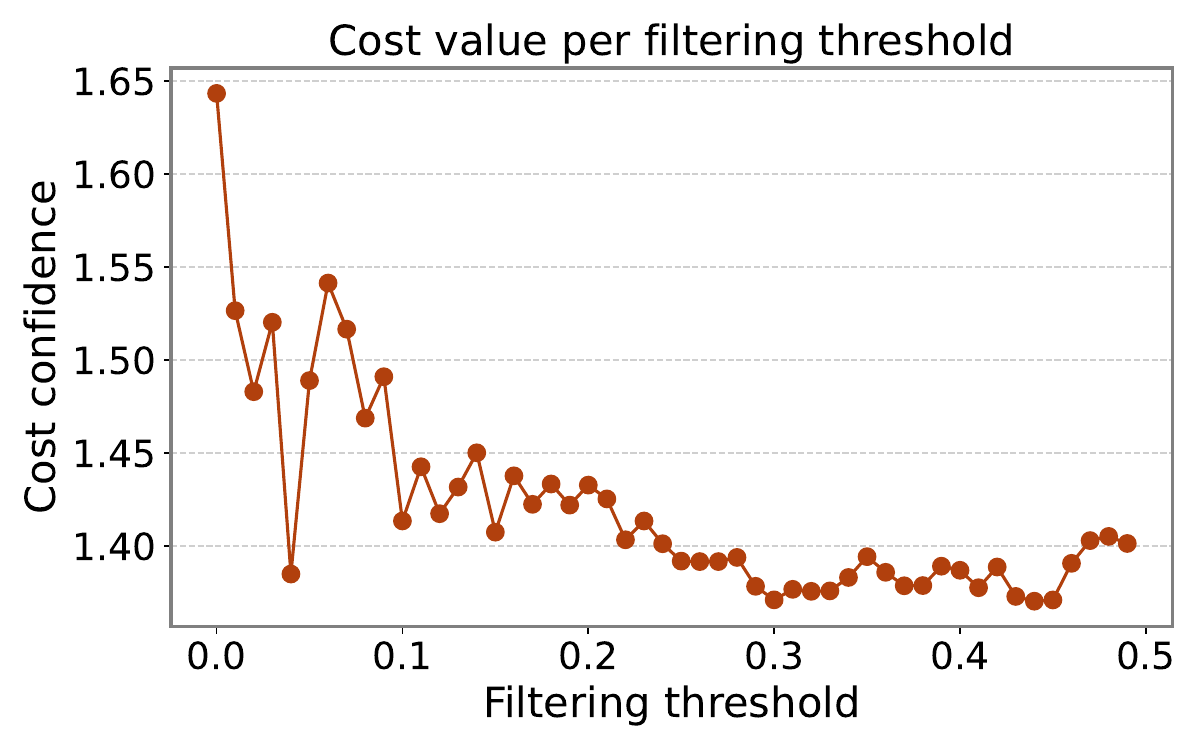}
        \caption*{(c) Brute force search.}
        \label{fig:image3}
    \end{minipage}
    \hfill
    \begin{minipage}{0.495\textwidth}
        \centering
        \includegraphics[width=\textwidth]{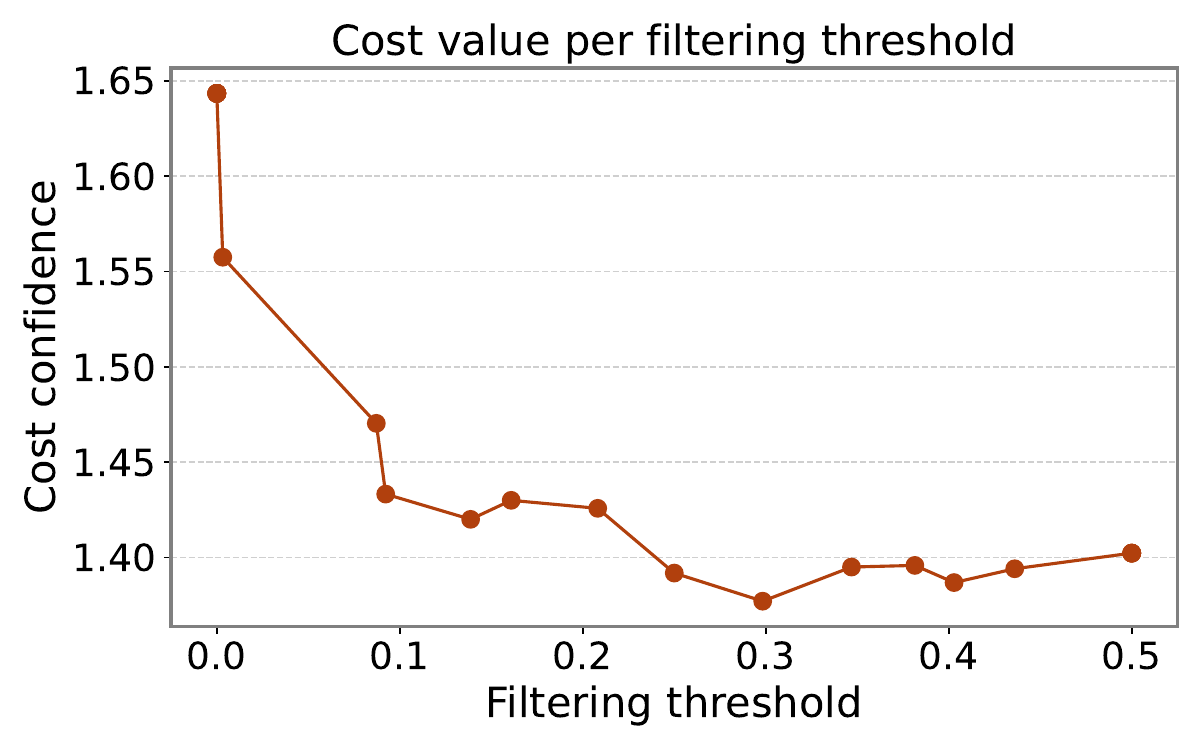}
        \caption*{(d) Simulated annealing search.}
        \label{fig:image4}
    \end{minipage}

    \caption{\small Cost values per filtering threshold applied tot he \textit{severity\_hsl} datset: a comparison among different strategies to minimize the cost function. Instances were filtered based on IH values, and confidence-based rejection was applied. For each filtering threshold we selected the rejection threshold that minimizes the cost. Values are averages over five different train-validation splits.  Figure (a) shows the result of a grid search trying specified values equally distributed in the interval being investigated. In Figure (b), after the grid search, a heuristic is adopted to explore the neighborhood of the filtering threshold that returned the smallest cost. Figure (c) presents the values obtained from a brute-force search, which tested 50 candidates for the filtering threshold. Finally, Figure (d) showcases the thresholds tested in a simulated annealing search, with 25 iterations. The smallest cost was achieved among all strategies in the second configuration, which combined the grid search and heuristic strategy. } \label{cost_optimization}
\end{figure*}

Three approaches were adopted to determine the filtering threshold \(T_f\). First, a grid search was conducted, followed by the heuristic step already described in Section \ref{sec_heuristic}. It begins with a grid search over five predefined values of \(T_f\). Starting with the threshold that yields the smallest cost, an exploration is performed in both directions, searching its neighborhood to refine the threshold and locate a better value that minimizes the cost function. The complete procedure is outlined in  Algorithm \ref{alg_heuristic}. Figure \ref{fig:comparison}(a)-(b) presents the result of both grid and heuristic search. It is replicated here on Figure \ref{cost_optimization} to facilitate direct comparison among the three optimization methods.

We then conducted a brute force search over 50 candidate values for the filtering threshold, ranging from 0 to 0.5 in increments of 0.01. The results of this investigation are presented in Figure \ref{cost_optimization}. When analyzing the brute-force search, we observe exponential behavior initially as we remove more instances. Subsequently, the curve begins to oscillate within an interval. This oscillation may reflect a trade-off between eliminating problematic instances and preserving valuable information. 

We also implemented a simulated annealing search. Simulated annealing is a metaheuristic inspired by the annealing process in metallurgy, in which a material is heated and then slowly cooled to reduce defects and find a stable configuration. 
The algorithm starts with an initial filtering threshold \(T_{f_0} = 0.25\) and an initial temperature \(T_0 = 100\). In each iteration, a new filtering threshold \(T_f'\) is generated by adding a random value in the interval \([-0.25, 0.25]\) to the current \(T_f\).

The change in cost is calculated as \( \Delta \text{Cost} = \text{Cost}(T_f') - \text{Cost}(T_f) \), and the new threshold is accepted if it satisfies one of the following criteria:

\begin{itemize}
    \item If \( \Delta \text{Cost} < 0 \) (i.e., the new solution is better), accept it.
    \item If \( \Delta \text{Cost} > 0 \), accept \( T_f' \) with probability \( p = \exp(-\Delta \text{Cost} / T) \).
\end{itemize}

This probabilistic acceptance allows the algorithm to escape local minima by occasionally accepting worse solutions. At each iteration, the temperature is decreased by multiplying it by 0.95 and making the search more conservative. The algorithm runs for 25 iterations. The simulated annealing procedure is described in Algorithm~\ref{alg_sa}.

In theory, simulated annealing can be guaranteed to find the global optimum under very strict conditions: temperature must decrease infinitely slowly and the algorithm must be run for an infinite number of steps at each temperature level \cite{kirkpatrick1983optimization}. Although these ideal conditions are unfeasible, in practice, simulated annealing often finds a good approximation to the global optimum, especially in large, complex search spaces with many local optimums \cite{siddique2016simulated}. 

For the dataset under study, all approaches find the minimum cost around \(T_f\) = 0.3. However, when balancing strategies and considering the computation cost, the economy of the heuristic method justifies its adoption. 

Although this result is based on a single dataset and may not be generalizable to other scenarios, it provides empirical evidence that the heuristic is a reasonable and effective strategy. Furthermore, its low computational cost supports its adoption over more resource-intensive methods.

We do not claim that the heuristic will yield the optimal solution. 
However, despite its simplicity, it offers a systematic and reasonable search of the cost function. It can be applied without negative consequences, 
because it guarantees that it will never return a result worse than the best value found in the initial grid search.

\begin{algorithm}    [h!]
\caption{Simulated annealing algorithm} \label{alg_sa}

\textbf{Definitions:} \\
\hangindent=1cm \hangafter=1
\(T_{f}\):\quad\hspace{0.5cm} filtering threshold.\\
\(Temp\):\quad\hspace{0.01cm} temperature. \\
\(T_{down}\):\quad\hspace{0.01cm} cooling rate set to 0.95. \\
\(iter_n\):\quad\hspace{0.15cm} number of iterations set to 25.\\
cost\((T_{f})\): best cost for \(T_{f}\) obtained selecting the best rejecting threshold \(T_{r}\). \\ 
\(bond_l\):\quad\hspace{0.1cm} lower bound set to 0.5.  \\
\(bond_u\):\quad\hspace{0.01cm} upper bound set to 1.  

\noindent\rule{\textwidth}{0.4pt} 
\begin{algorithmic}[1]
\State Initialize features: \(T_f \gets 0.25  \) and \( Temp \gets 100  \)
\State Train a new model using the filtered data and evaluate it on the validation set.
\State Measure \(cost(T_{f})\).
\For{$i = 0$ \textbf{to} $iter\_n$} 

\State Generate a new $T_{f'}$ value by adding a random perturbation to the current $T_f$
\[
T_{f'} \gets T_f + \delta, \quad \text{where } \delta \sim \mathcal{U}(-0.25, 0.25)
\]\[
T_{f'} \gets \max\left(\min\left(T_{f'}, bond_u \right), bond_l\right)
\]
\State Filter the training data using \(T_f'\).
\State Train a new model using the filtered data and evaluate it on the validation set.
\State Measure \(cost(T_{f'})\).
\State Calculate the cost variation: \( \Delta \gets \text{cost}(T_f') -\text{cost}(T_f) \)
\If{$\Delta  < 0$ \textbf{or} $\exp\left(-\Delta / temp\right) > \mathcal{U}(0,1)$}
\State Update:  \(T_f \gets T_f'\) and \text{cost}\((T_f) \gets \text{cost}(T_f') \)
\EndIf
\State Cool down the temperature: \( Temp \gets Temp * T_{down} \)
\EndFor
\State \Return \(T_f\)

\end{algorithmic}
\end{algorithm}

\begin{algorithm}   [h!] 
\caption{Heuristic optimization algorithm} \label{alg_heuristic}
\textbf{Definitions:} \\
\hangindent=1cm \hangafter=1
\(T_{f}\): \quad\quad\hspace{0.1cm} best filtering threshold resulted from the grid search. \\
\(\text{cost}(T_{f})\):\hspace{0.05cm} best cost for \(T_{f}\) obtained selecting the best rejecting threshold \(T_{r}\). \\ 
\(T_{f'}\): \quad\quad\hspace{0.05cm} a neighboring $T_f$ value, candidate to have a lower cost than $T_f$. \\
\(\text{cost}(T_{f'})\): the cost of using \(T_{f'}\). 

\noindent\rule{\textwidth}{0.4pt} 
\begin{algorithmic}[1]

\While{$|\text{cost}(T_{f}) - \text{cost}(T_{f'})| \geq \delta$ \textbf{ and } $\frac{T_{f} + T_{f'}}{2} \geq \epsilon_0$} \vspace{0.3cm}
    \State Define a new filtering threshold: \(T_{f''} = \frac{T_{f} + T_{f'}}{2} \)
    \State Train a new model using the filtered data and evaluate it on the validation set.
    \State Measure \(cost(T_{f''})\).    
    \State Update the thresholds: 
    \( 
    T_f \gets \argmin\limits_{x \in \{T_f,\ T_{f''}\}} \text{cost}(x)
    \), \quad
    \( 
    T_{f'} \gets \argmax\limits_{x \in \{T_f,\ T_{f''}\}} \text{cost}(x)
    \)
\EndWhile
\State \Return $T_f$ and $\text{cost}(T_f)$
\end{algorithmic}
\end{algorithm}

\clearpage

\

\section{Additional information} \label{apd}

\textcolor{black}{Table \ref{main_xgb} describes the hyperparameters adopted in the XGBoost model when trained as the main model of the framework. 
}
\begin{table}[h!]
\caption{\small XGBoost hyperparameters used for the main model.} 
\begin{tabular}{ll}

\toprule
\textbf{Parameter}            & \textbf{Value}          \\ \midrule
\textit{learning\_rate}       & 0.01                    \\ 
\textit{n\_estimators}        & 1000                    \\ 
\textit{max\_depth}           & 8                       \\ 
\textit{min\_child\_weight}   & 1                       \\
\textit{gamma}                & 0                       \\ 
\textit{subsample}            & 0.8                     \\ 
\textit{colsample\_bytree}    & 0.8                     \\ 
\textit{objective}            & 'binary:logistic'       \\ 
\textit{eval\_metric}         & 'logloss'               \\ 
\textit{scale\_pos\_weight}   & 1                       \\ 
\textit{seed}                 & 27                      \\ \bottomrule
\end{tabular}
\label{main_xgb}
\end{table}

\end{appendices}


\end{document}

%% file: sn-article.bbl

%% file: sn-article.bbl
\begin{thebibliography}{52}
\ifx \bisbn   \undefined \def \bisbn  #1{ISBN #1}\fi
\ifx \binits  \undefined \def \binits#1{#1}\fi
\ifx \bauthor  \undefined \def \bauthor#1{#1}\fi
\ifx \batitle  \undefined \def \batitle#1{#1}\fi
\ifx \bjtitle  \undefined \def \bjtitle#1{#1}\fi
\ifx \bvolume  \undefined \def \bvolume#1{\textbf{#1}}\fi
\ifx \byear  \undefined \def \byear#1{#1}\fi
\ifx \bissue  \undefined \def \bissue#1{#1}\fi
\ifx \bfpage  \undefined \def \bfpage#1{#1}\fi
\ifx \blpage  \undefined \def \blpage #1{#1}\fi
\ifx \burl  \undefined \def \burl#1{\textsf{#1}}\fi
\ifx \doiurl  \undefined \def \doiurl#1{\url{https://doi.org/#1}}\fi
\ifx \betal  \undefined \def \betal{\textit{et al.}}\fi
\ifx \binstitute  \undefined \def \binstitute#1{#1}\fi
\ifx \binstitutionaled  \undefined \def \binstitutionaled#1{#1}\fi
\ifx \bctitle  \undefined \def \bctitle#1{#1}\fi
\ifx \beditor  \undefined \def \beditor#1{#1}\fi
\ifx \bpublisher  \undefined \def \bpublisher#1{#1}\fi
\ifx \bbtitle  \undefined \def \bbtitle#1{#1}\fi
\ifx \bedition  \undefined \def \bedition#1{#1}\fi
\ifx \bseriesno  \undefined \def \bseriesno#1{#1}\fi
\ifx \blocation  \undefined \def \blocation#1{#1}\fi
\ifx \bsertitle  \undefined \def \bsertitle#1{#1}\fi
\ifx \bsnm \undefined \def \bsnm#1{#1}\fi
\ifx \bsuffix \undefined \def \bsuffix#1{#1}\fi
\ifx \bparticle \undefined \def \bparticle#1{#1}\fi
\ifx \barticle \undefined \def \barticle#1{#1}\fi
\bibcommenthead
\ifx \bconfdate \undefined \def \bconfdate #1{#1}\fi
\ifx \botherref \undefined \def \botherref #1{#1}\fi
\ifx \url \undefined \def \url#1{\textsf{#1}}\fi
\ifx \bchapter \undefined \def \bchapter#1{#1}\fi
\ifx \bbook \undefined \def \bbook#1{#1}\fi
\ifx \bcomment \undefined \def \bcomment#1{#1}\fi
\ifx \oauthor \undefined \def \oauthor#1{#1}\fi
\ifx \citeauthoryear \undefined \def \citeauthoryear#1{#1}\fi
\ifx \endbibitem  \undefined \def \endbibitem {}\fi
\ifx \bconflocation  \undefined \def \bconflocation#1{#1}\fi
\ifx \arxivurl  \undefined \def \arxivurl#1{\textsf{#1}}\fi
\csname PreBibitemsHook\endcsname

\bibitem[\protect\citeauthoryear{H{\"u}llermeier and Waegeman}{2021}]{hullermeier2021aleatoric}
\begin{barticle}
\bauthor{\bsnm{H{\"u}llermeier}, \binits{E.}},
\bauthor{\bsnm{Waegeman}, \binits{W.}}:
\batitle{Aleatoric and epistemic uncertainty in machine learning: An introduction to concepts and methods}.
\bjtitle{Machine learning}
\bvolume{110}(\bissue{3}),
\bfpage{457}--\blpage{506}
(\byear{2021})
\end{barticle}
\endbibitem

\bibitem[\protect\citeauthoryear{Hendrickx et~al.}{2024}]{hendrickx2024machine}
\begin{botherref}
\oauthor{\bsnm{Hendrickx}, \binits{K.}},
\oauthor{\bsnm{Perini}, \binits{L.}},
\oauthor{\bsnm{Plas}, \binits{D.}},
\oauthor{\bsnm{Meert}, \binits{W.}},
\oauthor{\bsnm{Davis}, \binits{J.}}:
Machine learning with a reject option: A survey.
Machine Learning,
1--38
(2024)
\end{botherref}
\endbibitem

\bibitem[\protect\citeauthoryear{Smith et~al.}{2014}]{smith2014instance}
\begin{barticle}
\bauthor{\bsnm{Smith}, \binits{M.R.}},
\bauthor{\bsnm{Martinez}, \binits{T.}},
\bauthor{\bsnm{Giraud-Carrier}, \binits{C.}}:
\batitle{An instance level analysis of data complexity}.
\bjtitle{Machine learning}
\bvolume{95}(\bissue{2}),
\bfpage{225}--\blpage{256}
(\byear{2014})
\end{barticle}
\endbibitem

\bibitem[\protect\citeauthoryear{Anik and Bunt}{2021}]{anik2021data}
\begin{bchapter}
\bauthor{\bsnm{Anik}, \binits{A.I.}},
\bauthor{\bsnm{Bunt}, \binits{A.}}:
\bctitle{Data-centric explanations: Explaining training data of machine learning systems to promote transparency}.
In: \bbtitle{Proceedings of the 2021 CHI Conference on Human Factors in Computing Systems},
pp. \bfpage{1}--\blpage{13}
(\byear{2021})
\end{bchapter}
\endbibitem

\bibitem[\protect\citeauthoryear{Rodriguez-Morales et~al.}{2020}]{rodriguez2020covid}
\begin{barticle}
\bauthor{\bsnm{Rodriguez-Morales}, \binits{A.J.}},
\bauthor{\bsnm{Gallego}, \binits{V.}},
\bauthor{\bsnm{Escalera-Antezana}, \binits{J.P.}},
\bauthor{\bsnm{M{\'e}ndez}, \binits{C.A.}},
\bauthor{\bsnm{Zambrano}, \binits{L.I.}},
\bauthor{\bsnm{Franco-Paredes}, \binits{C.}},
\bauthor{\bsnm{Su{\'a}rez}, \binits{J.A.}},
\bauthor{\bsnm{Rodriguez-Enciso}, \binits{H.D.}},
\bauthor{\bsnm{Balbin-Ramon}, \binits{G.J.}},
\bauthor{\bsnm{Savio-Larriera}, \binits{E.}}, \betal:
\batitle{Covid-19 in latin america: The implications of the first confirmed case in brazil}.
\bjtitle{Travel medicine and infectious disease}
\bvolume{35},
\bfpage{101613}
(\byear{2020})
\end{barticle}
\endbibitem

\bibitem[\protect\citeauthoryear{Valeriano et~al.}{2022}]{valeriano2022let}
\begin{bchapter}
\bauthor{\bsnm{Valeriano}, \binits{M.G.}},
\bauthor{\bsnm{Kiffer}, \binits{C.R.}},
\bauthor{\bsnm{Higino}, \binits{G.}},
\bauthor{\bsnm{Zan{\~a}o}, \binits{P.}},
\bauthor{\bsnm{Barbosa}, \binits{D.A.}},
\bauthor{\bsnm{Moreira}, \binits{P.A.}},
\bauthor{\bsnm{Santos}, \binits{P.C.J.}},
\bauthor{\bsnm{Grinbaum}, \binits{R.}},
\bauthor{\bsnm{Lorena}, \binits{A.C.}}:
\bctitle{Let the data speak: analysing data from multiple health centers of the s{\~a}o paulo metropolitan area for covid-19 clinical deterioration prediction}.
In: \bbtitle{2022 22nd IEEE International Symposium on Cluster, Cloud and Internet Computing (CCGrid)},
pp. \bfpage{948}--\blpage{951}
(\byear{2022}).
\bcomment{IEEE}
\end{bchapter}
\endbibitem

\bibitem[\protect\citeauthoryear{Seedat et~al.}{2022}]{seedat2022dc}
\begin{botherref}
\oauthor{\bsnm{Seedat}, \binits{N.}},
\oauthor{\bsnm{Imrie}, \binits{F.}},
\oauthor{\bsnm{Schaar}, \binits{M.}}:
Dc-check: A data-centric ai checklist to guide the development of reliable machine learning systems.
arXiv preprint arXiv:2211.05764
(2022)
\end{botherref}
\endbibitem

\bibitem[\protect\citeauthoryear{Sambasivan et~al.}{2021}]{sambasivan2021everyone}
\begin{bchapter}
\bauthor{\bsnm{Sambasivan}, \binits{N.}},
\bauthor{\bsnm{Kapania}, \binits{S.}},
\bauthor{\bsnm{Highfill}, \binits{H.}},
\bauthor{\bsnm{Akrong}, \binits{D.}},
\bauthor{\bsnm{Paritosh}, \binits{P.}},
\bauthor{\bsnm{Aroyo}, \binits{L.M.}}:
\bctitle{“everyone wants to do the model work, not the data work”: Data cascades in high-stakes ai}.
In: \bbtitle{Proceedings of the 2021 CHI Conference on Human Factors in Computing Systems},
pp. \bfpage{1}--\blpage{15}
(\byear{2021})
\end{bchapter}
\endbibitem

\bibitem[\protect\citeauthoryear{Zha et~al.}{2025}]{zha2025data}
\begin{barticle}
\bauthor{\bsnm{Zha}, \binits{D.}},
\bauthor{\bsnm{Bhat}, \binits{Z.P.}},
\bauthor{\bsnm{Lai}, \binits{K.-H.}},
\bauthor{\bsnm{Yang}, \binits{F.}},
\bauthor{\bsnm{Jiang}, \binits{Z.}},
\bauthor{\bsnm{Zhong}, \binits{S.}},
\bauthor{\bsnm{Hu}, \binits{X.}}:
\batitle{Data-centric artificial intelligence: A survey}.
\bjtitle{ACM Computing Surveys}
\bvolume{57}(\bissue{5}),
\bfpage{1}--\blpage{42}
(\byear{2025})
\end{barticle}
\endbibitem

\bibitem[\protect\citeauthoryear{Seedat et~al.}{2022a}]{seedat2022dataiq}
\begin{barticle}
\bauthor{\bsnm{Seedat}, \binits{N.}},
\bauthor{\bsnm{Crabb{\'e}}, \binits{J.}},
\bauthor{\bsnm{Bica}, \binits{I.}},
\bauthor{\bsnm{Schaar}, \binits{M.}}:
\batitle{Data-iq: Characterizing subgroups with heterogeneous outcomes in tabular data}.
\bjtitle{Advances in Neural Information Processing Systems}
\bvolume{35},
\bfpage{23660}--\blpage{23674}
(\byear{2022})
\end{barticle}
\endbibitem

\bibitem[\protect\citeauthoryear{Seedat et~al.}{2022b}]{seedat2022data}
\begin{bchapter}
\bauthor{\bsnm{Seedat}, \binits{N.}},
\bauthor{\bsnm{Crabb{\'e}}, \binits{J.}},
\bauthor{\bsnm{Schaar}, \binits{M.}}:
\bctitle{Data-suite: Data-centric identification of in-distribution incongruous examples}.
In: \bbtitle{International Conference on Machine Learning},
pp. \bfpage{19467}--\blpage{19496}
(\byear{2022}).
\bcomment{PMLR}
\end{bchapter}
\endbibitem

\bibitem[\protect\citeauthoryear{Seedat et~al.}{}]{seedat2024dissecting}
\begin{botherref}
\oauthor{\bsnm{Seedat}, \binits{N.}},
\oauthor{\bsnm{Imrie}, \binits{F.}},
\oauthor{\bsnm{Schaar}, \binits{M.}}:
Dissecting sample hardness: A fine-grained analysis of hardness characterization methods for data-centric ai.
In: The Twelfth International Conference on Learning Representations
\end{botherref}
\endbibitem

\bibitem[\protect\citeauthoryear{Liu et~al.}{2024}]{liu2024instance}
\begin{barticle}
\bauthor{\bsnm{Liu}, \binits{C.}},
\bauthor{\bsnm{Smith-Miles}, \binits{K.}},
\bauthor{\bsnm{Wauters}, \binits{T.}},
\bauthor{\bsnm{Costa}, \binits{A.M.}}:
\batitle{Instance space analysis for 2d bin packing mathematical models}.
\bjtitle{European Journal of Operational Research}
\bvolume{315}(\bissue{2}),
\bfpage{484}--\blpage{498}
(\byear{2024})
\end{barticle}
\endbibitem

\bibitem[\protect\citeauthoryear{Paul et~al.}{2021}]{paul2021deep}
\begin{barticle}
\bauthor{\bsnm{Paul}, \binits{M.}},
\bauthor{\bsnm{Ganguli}, \binits{S.}},
\bauthor{\bsnm{Dziugaite}, \binits{G.K.}}:
\batitle{Deep learning on a data diet: Finding important examples early in training}.
\bjtitle{Advances in neural information processing systems}
\bvolume{34},
\bfpage{20596}--\blpage{20607}
(\byear{2021})
\end{barticle}
\endbibitem

\bibitem[\protect\citeauthoryear{Valeriano et~al.}{2024a}]{valeriano2024explaining}
\begin{bchapter}
\bauthor{\bsnm{Valeriano}, \binits{M.G.}},
\bauthor{\bsnm{Pereira}, \binits{J.L.J.}},
\bauthor{\bsnm{Veiga~Kiffer}, \binits{C.R.}},
\bauthor{\bsnm{Lorena}, \binits{A.C.}}:
\bctitle{Explaining instances in the health domain based on the exploration of a dataset's hardness embedding}.
In: \bbtitle{Proceedings of the Genetic and Evolutionary Computation Conference Companion},
pp. \bfpage{1598}--\blpage{1606}
(\byear{2024})
\end{bchapter}
\endbibitem

\bibitem[\protect\citeauthoryear{Valeriano et~al.}{2024b}]{valeriano2024improving}
\begin{bchapter}
\bauthor{\bsnm{Valeriano}, \binits{M.}},
\bauthor{\bsnm{Kiffer}, \binits{C.}},
\bauthor{\bsnm{Lorena}, \binits{A.}}:
\bctitle{Improving models performance in a data-centric approach applied to the healthcare domain}.
In: \bbtitle{Symposium on Knowledge Discovery, Mining and Learning (KDMiLe)},
pp. \bfpage{57}--\blpage{64}
(\byear{2024}).
\bcomment{SBC}
\end{bchapter}
\endbibitem

\bibitem[\protect\citeauthoryear{Koh and Liang}{2017}]{Koh2017}
\begin{botherref}
\oauthor{\bsnm{Koh}, \binits{P.W.}},
\oauthor{\bsnm{Liang}, \binits{P.}}:
{Understanding Black-box Predictions via Influence Functions}.
Proceedings of the 34th International Conference on Machine Learning,
1885--1894
(2017)
\end{botherref}
\endbibitem

\bibitem[\protect\citeauthoryear{Valeriano et~al.}{2025}]{Valeriano2025Beyond}
\begin{botherref}
\oauthor{\bsnm{Valeriano}, \binits{M.G.}},
\oauthor{\bsnm{Kiffer}, \binits{C.R.V.}},
\oauthor{\bsnm{Lorena}, \binits{A.C.}}:
Beyond Filtering: Leveraging Instance Hardness for Data-Centric Machine Learning in Healthcare.
To appear in the Proceedings of the International Joint Conference on Neural Networks (IJCNN), 2025.
(2025)
\end{botherref}
\endbibitem

\bibitem[\protect\citeauthoryear{Traub et~al.}{2024}]{traub2024overcoming}
\begin{barticle}
\bauthor{\bsnm{Traub}, \binits{J.}},
\bauthor{\bsnm{Bungert}, \binits{T.J.}},
\bauthor{\bsnm{L{\"u}th}, \binits{C.T.}},
\bauthor{\bsnm{Baumgartner}, \binits{M.}},
\bauthor{\bsnm{Maier-Hein}, \binits{K.H.}},
\bauthor{\bsnm{Maier-Hein}, \binits{L.}},
\bauthor{\bsnm{J{\"a}ger}, \binits{P.F.}}:
\batitle{Overcoming common flaws in the evaluation of selective classification systems}.
\bjtitle{Advances in Neural Information Processing Systems}
\bvolume{37},
\bfpage{2323}--\blpage{2347}
(\byear{2024})
\end{barticle}
\endbibitem

\bibitem[\protect\citeauthoryear{Bartlett and Wegkamp}{2008}]{bartlett2008classification}
\begin{botherref}
\oauthor{\bsnm{Bartlett}, \binits{P.L.}},
\oauthor{\bsnm{Wegkamp}, \binits{M.H.}}:
Classification with a reject option using a hinge loss.
Journal of Machine Learning Research
\textbf{9}(8)
(2008)
\end{botherref}
\endbibitem

\bibitem[\protect\citeauthoryear{Varshney}{2011}]{varshney2011risk}
\begin{bchapter}
\bauthor{\bsnm{Varshney}, \binits{K.R.}}:
\bctitle{A risk bound for ensemble classification with a reject option}.
In: \bbtitle{2011 IEEE Statistical Signal Processing Workshop (SSP)},
pp. \bfpage{769}--\blpage{772}
(\byear{2011}).
\bcomment{IEEE}
\end{bchapter}
\endbibitem

\bibitem[\protect\citeauthoryear{Xu and Chetia}{2023}]{xu2023efficient}
\begin{bchapter}
\bauthor{\bsnm{Xu}, \binits{H.}},
\bauthor{\bsnm{Chetia}, \binits{C.}}:
\bctitle{An efficient selective ensemble learning with rejection approach for classification}.
In: \bbtitle{Proceedings of the 32nd ACM International Conference on Information and Knowledge Management},
pp. \bfpage{2816}--\blpage{2825}
(\byear{2023})
\end{bchapter}
\endbibitem

\bibitem[\protect\citeauthoryear{Lakshminarayanan et~al.}{2017}]{lakshminarayanan2017simple}
\begin{botherref}
\oauthor{\bsnm{Lakshminarayanan}, \binits{B.}},
\oauthor{\bsnm{Pritzel}, \binits{A.}},
\oauthor{\bsnm{Blundell}, \binits{C.}}:
Simple and scalable predictive uncertainty estimation using deep ensembles.
Advances in neural information processing systems
\textbf{30}
(2017)
\end{botherref}
\endbibitem

\bibitem[\protect\citeauthoryear{B{\"o}ken}{2021}]{boken2021appropriateness}
\begin{barticle}
\bauthor{\bsnm{B{\"o}ken}, \binits{B.}}:
\batitle{On the appropriateness of platt scaling in classifier calibration}.
\bjtitle{Information Systems}
\bvolume{95},
\bfpage{101641}
(\byear{2021})
\end{barticle}
\endbibitem

\bibitem[\protect\citeauthoryear{Collins et~al.}{2015}]{collins2015transparent}
\begin{barticle}
\bauthor{\bsnm{Collins}, \binits{G.S.}},
\bauthor{\bsnm{Reitsma}, \binits{J.B.}},
\bauthor{\bsnm{Altman}, \binits{D.G.}},
\bauthor{\bsnm{Moons}, \binits{K.G.}}:
\batitle{Transparent reporting of a multivariable prediction model for individual prognosis or diagnosis (tripod) the tripod statement}.
\bjtitle{Circulation}
\bvolume{131}(\bissue{2}),
\bfpage{211}--\blpage{219}
(\byear{2015})
\end{barticle}
\endbibitem

\bibitem[\protect\citeauthoryear{Paiva et~al.}{2022}]{paiva2022relating}
\begin{botherref}
\oauthor{\bsnm{Paiva}, \binits{P.Y.A.}},
\oauthor{\bsnm{Moreno}, \binits{C.C.}},
\oauthor{\bsnm{Smith-Miles}, \binits{K.}},
\oauthor{\bsnm{Valeriano}, \binits{M.G.}},
\oauthor{\bsnm{Lorena}, \binits{A.C.}}:
Relating instance hardness to classification performance in a dataset: a visual approach.
Machine Learning,
1--39
(2022)
\end{botherref}
\endbibitem

\bibitem[\protect\citeauthoryear{Paiva et~al.}{2021}]{paiva2021pyhard}
\begin{bchapter}
\bauthor{\bsnm{Paiva}, \binits{P.Y.A.}},
\bauthor{\bsnm{Smith-Miles}, \binits{K.}},
\bauthor{\bsnm{Valeriano}, \binits{M.G.}},
\bauthor{\bsnm{Lorena}, \binits{A.C.}}:
\bctitle{Pyhard: A novel tool for generating hardness embeddings to support data-centric analysis}.
In: \bbtitle{Proceedings of the 35th Conference on Neural Information Processing Systems (NeurIPS 2021) Workshop on Data-Centric AI (DCAI)}
(\byear{2021})
\end{bchapter}
\endbibitem

\bibitem[\protect\citeauthoryear{Batista et~al.}{2004}]{batista2004study}
\begin{barticle}
\bauthor{\bsnm{Batista}, \binits{G.E.}},
\bauthor{\bsnm{Prati}, \binits{R.C.}},
\bauthor{\bsnm{Monard}, \binits{M.C.}}:
\batitle{A study of the behavior of several methods for balancing machine learning training data}.
\bjtitle{ACM SIGKDD explorations newsletter}
\bvolume{6}(\bissue{1}),
\bfpage{20}--\blpage{29}
(\byear{2004})
\end{barticle}
\endbibitem

\bibitem[\protect\citeauthoryear{Chen and Guestrin}{2016}]{chen2016xgboost}
\begin{bchapter}
\bauthor{\bsnm{Chen}, \binits{T.}},
\bauthor{\bsnm{Guestrin}, \binits{C.}}:
\bctitle{Xgboost: A scalable tree boosting system}.
In: \bbtitle{Proceedings of the 22nd Acm Sigkdd International Conference on Knowledge Discovery and Data Mining},
pp. \bfpage{785}--\blpage{794}
(\byear{2016})
\end{bchapter}
\endbibitem

\bibitem[\protect\citeauthoryear{Barcellos et~al.}{2024}]{barcellos2024climate}
\begin{barticle}
\bauthor{\bsnm{Barcellos}, \binits{C.}},
\bauthor{\bsnm{Matos}, \binits{V.}},
\bauthor{\bsnm{Lana}, \binits{R.M.}},
\bauthor{\bsnm{Lowe}, \binits{R.}}:
\batitle{Climate change, thermal anomalies, and the recent progression of dengue in brazil}.
\bjtitle{Scientific reports}
\bvolume{14}(\bissue{1}),
\bfpage{5948}
(\byear{2024})
\end{barticle}
\endbibitem

\bibitem[\protect\citeauthoryear{Gon{\c{c}}alves et~al.}{2017}]{gonccalves2017expanding}
\begin{barticle}
\bauthor{\bsnm{Gon{\c{c}}alves}, \binits{M.}},
\bauthor{\bsnm{Umpierre}, \binits{R.}},
\bauthor{\bsnm{D’Avila}, \binits{O.}},
\bauthor{\bsnm{Katz}, \binits{N.}},
\bauthor{\bsnm{Mengue}, \binits{S.}},
\bauthor{\bsnm{Siqueira}, \binits{A.}},
\bauthor{\bsnm{Carrard}, \binits{V.}},
\bauthor{\bsnm{Schmitz}, \binits{C.}},
\bauthor{\bsnm{Molina-Bastos}, \binits{C.}},
\bauthor{\bsnm{Rados}, \binits{D.}}, \betal:
\batitle{Expanding primary care access: a telehealth success story}.
\bjtitle{Annals of Family Medicine}
\bvolume{15}(\bissue{4}),
\bfpage{383}
(\byear{2017})
\end{barticle}
\endbibitem

\bibitem[\protect\citeauthoryear{Souza et~al.}{2020}]{souza2020bertimbau}
\begin{bchapter}
\bauthor{\bsnm{Souza}, \binits{F.}},
\bauthor{\bsnm{Nogueira}, \binits{R.}},
\bauthor{\bsnm{Lotufo}, \binits{R.}}:
\bctitle{{BERT}imbau: pretrained {BERT} models for {B}razilian {P}ortuguese}.
In: \bbtitle{9th Brazilian Conference on Intelligent Systems, {BRACIS}, Rio Grande do Sul, Brazil, October 20-23 (to Appear)}
(\byear{2020})
\end{bchapter}
\endbibitem

\bibitem[\protect\citeauthoryear{Mello et~al.}{2020}]{mello2020opening}
\begin{botherref}
\oauthor{\bsnm{Mello}, \binits{L.E.}},
\oauthor{\bsnm{Suman}, \binits{A.}},
\oauthor{\bsnm{Medeiros}, \binits{C.B.}},
\oauthor{\bsnm{Prado}, \binits{C.A.}},
\oauthor{\bsnm{Rizzatti}, \binits{E.G.}},
\oauthor{\bsnm{Nunes}, \binits{F.L.}},
\oauthor{\bsnm{Barnab{\'e}}, \binits{G.F.}},
\oauthor{\bsnm{Ferreira}, \binits{J.E.}},
\oauthor{\bsnm{S{\'a}}, \binits{J.}},
\oauthor{\bsnm{Reis}, \binits{L.F.}}, et al.:
Opening brazilian covid-19 patient data to support world research on pandemics.
Zenodo
(2020)
\end{botherref}
\endbibitem

\bibitem[\protect\citeauthoryear{Wu et~al.}{2020}]{Wu2020}
\begin{barticle}
\bauthor{\bsnm{Wu}, \binits{Y.}},
\bauthor{\bsnm{Hou}, \binits{B.}},
\bauthor{\bsnm{Liu}, \binits{J.}},
\bauthor{\bsnm{Chen}, \binits{Y.}},
\bauthor{\bsnm{Zhong}, \binits{P.}}:
\batitle{Risk factors associated with long-term hospitalization in patients with {COVID}-19: A single-centered, retrospective study}.
\bjtitle{Frontiers in Medicine}
\bvolume{7}(\bissue{315}),
\bfpage{1}--\blpage{10}
(\byear{2020})
\end{barticle}
\endbibitem

\bibitem[\protect\citeauthoryear{Valeriano et~al.}{2024}]{valeriano2024understanding}
\begin{barticle}
\bauthor{\bsnm{Valeriano}, \binits{M.G.}},
\bauthor{\bsnm{Matran-Fernandez}, \binits{A.}},
\bauthor{\bsnm{Kiffer}, \binits{C.}},
\bauthor{\bsnm{Lorena}, \binits{A.C.}}:
\batitle{Understanding the performance of machine learning models from data-to patient-level}.
\bjtitle{ACM Journal of Data and Information Quality}
\bvolume{16}(\bissue{4}),
\bfpage{1}--\blpage{19}
(\byear{2024})
\end{barticle}
\endbibitem

\bibitem[\protect\citeauthoryear{McElfresh et~al.}{2023}]{mcelfresh2023neural}
\begin{barticle}
\bauthor{\bsnm{McElfresh}, \binits{D.}},
\bauthor{\bsnm{Khandagale}, \binits{S.}},
\bauthor{\bsnm{Valverde}, \binits{J.}},
\bauthor{\bsnm{Prasad~C}, \binits{V.}},
\bauthor{\bsnm{Ramakrishnan}, \binits{G.}},
\bauthor{\bsnm{Goldblum}, \binits{M.}},
\bauthor{\bsnm{White}, \binits{C.}}:
\batitle{When do neural nets outperform boosted trees on tabular data?}
\bjtitle{Advances in Neural Information Processing Systems}
\bvolume{36},
\bfpage{76336}--\blpage{76369}
(\byear{2023})
\end{barticle}
\endbibitem

\bibitem[\protect\citeauthoryear{Nunes et~al.}{2021}]{nunes2021using}
\begin{bchapter}
\bauthor{\bsnm{Nunes}, \binits{G.H.}},
\bauthor{\bsnm{Martins}, \binits{G.O.}},
\bauthor{\bsnm{Forster}, \binits{C.H.}},
\bauthor{\bsnm{Lorena}, \binits{A.C.}}:
\bctitle{Using instance hardness measures in curriculum learning}.
In: \bbtitle{Encontro Nacional de Intelig{\^e}ncia Artificial e Computacional (ENIAC)},
pp. \bfpage{177}--\blpage{188}
(\byear{2021}).
\bcomment{SBC}
\end{bchapter}
\endbibitem

\bibitem[\protect\citeauthoryear{Ferreira et~al.}{2024}]{ferreira2024measuring}
\begin{bchapter}
\bauthor{\bsnm{Ferreira}, \binits{E.V.}},
\bauthor{\bsnm{Prud{\^e}ncio}, \binits{R.B.C.}},
\bauthor{\bsnm{Lorena}, \binits{A.C.}}:
\bctitle{Measuring latent traits of instance hardness and classifier ability using boltzmann machines}.
In: \bbtitle{2024 International Joint Conference on Neural Networks (IJCNN)},
pp. \bfpage{1}--\blpage{8}
(\byear{2024}).
\bcomment{IEEE}
\end{bchapter}
\endbibitem

\bibitem[\protect\citeauthoryear{Kamiran et~al.}{2018}]{kamiran2018exploiting}
\begin{barticle}
\bauthor{\bsnm{Kamiran}, \binits{F.}},
\bauthor{\bsnm{Mansha}, \binits{S.}},
\bauthor{\bsnm{Karim}, \binits{A.}},
\bauthor{\bsnm{Zhang}, \binits{X.}}:
\batitle{Exploiting reject option in classification for social discrimination control}.
\bjtitle{Information Sciences}
\bvolume{425},
\bfpage{18}--\blpage{33}
(\byear{2018})
\end{barticle}
\endbibitem

\bibitem[\protect\citeauthoryear{Torquette et~al.}{2022}]{torquette2022characterizing}
\begin{bchapter}
\bauthor{\bsnm{Torquette}, \binits{G.P.}},
\bauthor{\bsnm{Nunes}, \binits{V.S.}},
\bauthor{\bsnm{Paiva}, \binits{P.Y.}},
\bauthor{\bsnm{Neto}, \binits{L.B.C.}},
\bauthor{\bsnm{Lorena}, \binits{A.C.}}:
\bctitle{Characterizing instance hardness in classification and regression problems}.
In: \bbtitle{Symposium on Knowledge Discovery, Mining and Learning (KDMiLe)},
pp. \bfpage{178}--\blpage{185}
(\byear{2022}).
\bcomment{SBC}
\end{bchapter}
\endbibitem

\bibitem[\protect\citeauthoryear{Sluijterman et~al.}{2024}]{sluijterman2024evaluate}
\begin{barticle}
\bauthor{\bsnm{Sluijterman}, \binits{L.}},
\bauthor{\bsnm{Cator}, \binits{E.}},
\bauthor{\bsnm{Heskes}, \binits{T.}}:
\batitle{How to evaluate uncertainty estimates in machine learning for regression?}
\bjtitle{Neural Networks}
\bvolume{173},
\bfpage{106203}
(\byear{2024})
\end{barticle}
\endbibitem

\bibitem[\protect\citeauthoryear{Xia et~al.}{2011}]{xia2011accurate}
\begin{barticle}
\bauthor{\bsnm{Xia}, \binits{L.C.}},
\bauthor{\bsnm{Cram}, \binits{J.A.}},
\bauthor{\bsnm{Chen}, \binits{T.}},
\bauthor{\bsnm{Fuhrman}, \binits{J.A.}},
\bauthor{\bsnm{Sun}, \binits{F.}}:
\batitle{Accurate genome relative abundance estimation based on shotgun metagenomic reads}.
\bjtitle{PloS one}
\bvolume{6}(\bissue{12}),
\bfpage{27992}
(\byear{2011})
\end{barticle}
\endbibitem

\bibitem[\protect\citeauthoryear{Franc et~al.}{2024}]{franc2024scod}
\begin{bchapter}
\bauthor{\bsnm{Franc}, \binits{V.}},
\bauthor{\bsnm{Paplham}, \binits{J.}},
\bauthor{\bsnm{Prusa}, \binits{D.}}:
\bctitle{Scod: From heuristics to theory}.
In: \bbtitle{European Conference on Computer Vision},
pp. \bfpage{424}--\blpage{441}
(\byear{2024}).
\bcomment{Springer}
\end{bchapter}
\endbibitem

\bibitem[\protect\citeauthoryear{Franc et~al.}{2023}]{franc2023optimal}
\begin{barticle}
\bauthor{\bsnm{Franc}, \binits{V.}},
\bauthor{\bsnm{Prusa}, \binits{D.}},
\bauthor{\bsnm{Voracek}, \binits{V.}}:
\batitle{Optimal strategies for reject option classifiers}.
\bjtitle{Journal of Machine Learning Research}
\bvolume{24}(\bissue{11}),
\bfpage{1}--\blpage{49}
(\byear{2023})
\end{barticle}
\endbibitem

\bibitem[\protect\citeauthoryear{Vanschoren et~al.}{2014}]{vanschoren2014openml}
\begin{barticle}
\bauthor{\bsnm{Vanschoren}, \binits{J.}},
\bauthor{\bsnm{Van~Rijn}, \binits{J.N.}},
\bauthor{\bsnm{Bischl}, \binits{B.}},
\bauthor{\bsnm{Torgo}, \binits{L.}}:
\batitle{Openml: networked science in machine learning}.
\bjtitle{ACM SIGKDD Explorations Newsletter}
\bvolume{15}(\bissue{2}),
\bfpage{49}--\blpage{60}
(\byear{2014})
\end{barticle}
\endbibitem

\bibitem[\protect\citeauthoryear{Cestnik et~al.}{1987}]{cestnik1987}
\begin{bchapter}
\bauthor{\bsnm{Cestnik}, \binits{B.}},
\bauthor{\bsnm{Kononenko}, \binits{I.}},
\bauthor{\bsnm{Bratko}, \binits{I.}}:
\bctitle{A knowledge-elicitation tool for sophisticated users}.
In: \bbtitle{Proceedings of the 2nd European Conference on European Working Session on Learning EWSL},
vol. \bseriesno{87}
(\byear{1987})
\end{bchapter}
\endbibitem

\bibitem[\protect\citeauthoryear{Clark et~al.}{1987}]{clark1987}
\begin{bchapter}
\bauthor{\bsnm{Clark}, \binits{P.}},
\bauthor{\bsnm{Niblett}, \binits{T.}}, \betal:
\bctitle{Induction in noisy domains.}
In: \bbtitle{EWSL},
pp. \bfpage{11}--\blpage{30}
(\byear{1987})
\end{bchapter}
\endbibitem

\bibitem[\protect\citeauthoryear{Michalski et~al.}{1986}]{michalski1986}
\begin{bchapter}
\bauthor{\bsnm{Michalski}, \binits{R.}},
\bauthor{\bsnm{Mozeti\v{c}}, \binits{I.}},
\bauthor{\bsnm{Hong}, \binits{J.}},
\bauthor{\bsnm{Lavra\v{c}}, \binits{N.}}:
\bctitle{The multi-purpose incremental learning system aq15 and its testing applications to three medical domains}.
In: \bbtitle{Proceedings of the Fifth National Conference on Artificial Intelligence},
pp. \bfpage{1041}--\blpage{1045}.
\bpublisher{Morgan Kaufmann},
\blocation{Philadelphia, PA}
(\byear{1986})
\end{bchapter}
\endbibitem

\bibitem[\protect\citeauthoryear{Asuncion et~al.}{2007}]{asuncion2007uci}
\begin{botherref}
\oauthor{\bsnm{Asuncion}, \binits{A.}},
\oauthor{\bsnm{Newman}, \binits{D.}}, et al.:
UCI machine learning repository.
Irvine, CA, USA
(2007)
\end{botherref}
\endbibitem

\bibitem[\protect\citeauthoryear{Dem{\v{s}}ar}{2006}]{demvsar2006statistical}
\begin{barticle}
\bauthor{\bsnm{Dem{\v{s}}ar}, \binits{J.}}:
\batitle{Statistical comparisons of classifiers over multiple data sets}.
\bjtitle{Journal of Machine learning research}
\bvolume{7}(\bissue{Jan}),
\bfpage{1}--\blpage{30}
(\byear{2006})
\end{barticle}
\endbibitem

\bibitem[\protect\citeauthoryear{Kirkpatrick et~al.}{1983}]{kirkpatrick1983optimization}
\begin{barticle}
\bauthor{\bsnm{Kirkpatrick}, \binits{S.}},
\bauthor{\bsnm{Gelatt~Jr}, \binits{C.D.}},
\bauthor{\bsnm{Vecchi}, \binits{M.P.}}:
\batitle{Optimization by simulated annealing}.
\bjtitle{science}
\bvolume{220}(\bissue{4598}),
\bfpage{671}--\blpage{680}
(\byear{1983})
\end{barticle}
\endbibitem

\bibitem[\protect\citeauthoryear{Siddique and Adeli}{2016}]{siddique2016simulated}
\begin{barticle}
\bauthor{\bsnm{Siddique}, \binits{N.}},
\bauthor{\bsnm{Adeli}, \binits{H.}}:
\batitle{Simulated annealing, its variants and engineering applications}.
\bjtitle{International Journal on Artificial Intelligence Tools}
\bvolume{25}(\bissue{06}),
\bfpage{1630001}
(\byear{2016})
\end{barticle}
\endbibitem

\end{thebibliography}
